%% file: main.tex
\documentclass{article}
\usepackage[main, preprint]{neurips_2026}  

\usepackage[utf8]{inputenc}
\usepackage[T1]{fontenc}
\usepackage{hyperref}
\usepackage{url}
\usepackage{booktabs}
\usepackage{amsfonts}
\usepackage{amsmath}
\usepackage{amssymb}
\usepackage{nicefrac}
\usepackage{microtype}
\usepackage{xcolor}
\usepackage{graphicx}
\usepackage{algorithm}
\usepackage{algpseudocode}
\usepackage{tikz}
\usetikzlibrary{positioning, arrows.meta, fit, backgrounds}
\usepackage{makecell}
\usepackage{longtable}

\title{Reducing cross-sample prediction churn in scientific machine learning}

\author{%
  Gordan Prastalo$^{1,2}$ \\
  \texttt{mail@prastalog.com}
  \And
  Kevin Maik Jablonka$^{2,3,4,5}$ \\
  \texttt{mail@kjablonka.com}
}

\newcommand{\affiliations}{%
  \begin{minipage}{\textwidth}
    \centering\small\itshape
    $^{1}$Helmholtz-Zentrum Berlin f\"ur Materialien und Energie GmbH,
    Hahn-Meitner-Platz 1, 14109 Berlin, Germany \\
    $^{2}$HIPOLE Jena (Helmholtz Institute for Polymers in Energy Applications
    Jena), Lessingstrasse 12--14, 07743 Jena, Germany \\
    $^{3}$Laboratory of Organic and Macromolecular Chemistry (IOMC),
    Friedrich Schiller University Jena, Humboldtstrasse 10, 07743 Jena, Germany \\
    $^{4}$Center for Energy and Environmental Chemistry Jena (CEEC Jena),
    Friedrich Schiller University Jena, Philosophenweg 7a, 07743 Jena, Germany \\
    $^{5}$Jena Center for Soft Matter (JCSM), Friedrich Schiller University
    Jena, Philosophenweg 7, 07743 Jena, Germany
  \end{minipage}%
}

\input{sections/macros}

\begin{document}
\maketitle
\affiliations\vspace{0.6em}

\input{sections/abstract}
\input{sections/introduction}

\input{sections/related_work}

\input{sections/measurement}

\input{sections/magnitudes}
\input{sections/methods}
\input{sections/experiments}
\input{sections/scope}
\input{sections/discussion}

\section*{Acknowledgements}
K.M.J.'s work is supported by the Carl Zeiss Foundation.
G.P.'s work was supported by the HPC Gateway measure of the Helmholtz
Association.  The authors thank Marti\~no R\'ios-Garc\'ia for feedback
on an early draft of the manuscript.

\section*{Declaration of generative AI and AI-assisted technologies in the research and writing process}
In addition to using Anthropic’s Claude models to write this declaration,
we used Anthropic's Claude models and OpenAI's Codex as copilots
during code development and cluster experiment submission, and
Claude additionally to improve language and readability of the
manuscript.  After using these services we reviewed and edited the
content as needed and take full responsibility for the content of
the publication.

\section*{Code availability}
Code is available at \url{https://github.com/lamalab-org/data-invariance}.

\bibliographystyle{plainnat}
\bibliography{references}

\appendix
\begin{center}
  {\LARGE\bfseries Appendix\par}
\end{center}
\addcontentsline{toc}{section}{Appendix}
\vspace{0.5em}
\input{sections/appendix}

\end{document}

%% file: sections/macros.tex
\newcommand{\nDatasetsHeadline}{9}
\newcommand{\nDatasetsHeldout}{8}
\newcommand{\overlapPctMid}{40}
\newcommand{\overlapPctLo}{0}
\newcommand{\overlapPctHi}{100}
\newcommand{\preregFilterPP}{5}
\newcommand{\preregMinTestSize}{60}
\newcommand{\preregTolerance}{0.02}
\newcommand{\nSeeds}{10}
\newcommand{\nSeedPairs}{45}
\newcommand{\canonicalSeed}{99}
\newcommand{\reviewFraction}{30}
\newcommand{\topDecilePct}{10}
\newcommand{\accDegradeFlagPP}{5}
\newcommand{\nCanonicalSeeds}{3}
\newcommand{\preregLambda}{300}
\newcommand{\churnMin}{8.0}
\newcommand{\churnMax}{21.8}
\newcommand{\accDiffMin}{1.3}
\newcommand{\accDiffMax}{4.2}
\newcommand{\churnAccRatioMin}{3}
\newcommand{\churnAccRatioMax}{14}
\newcommand{\symKLDatasetSpread}{3}

\newcommand{\baceMeanAccuracyDiff}{1.8}
\newcommand{\bagFiveLow}{40}
\newcommand{\bagFiveHigh}{54}
\newcommand{\twinMedianReduction}{65}
\newcommand{\twinFurtherMedianReductionVsBagTwo}{45}
\newcommand{\paramSideLow}{-22.3}
\newcommand{\paramSideHigh}{+12.5}

\newcommand{\baceErmChurn}{16.1}
\newcommand{\baceTwinChurn}{5.7}
\newcommand{\deepEnsAccDelta}{+0.1}
\newcommand{\twinAccDelta}{-0.7}
\newcommand{\symKLBagFoldLow}{6}
\newcommand{\symKLBagFoldHigh}{9}
\newcommand{\symKLTwinAdditionalFoldMedian}{9}
\newcommand{\friedmanChi}{44.8}
\newcommand{\friedmanP}{5.2{\times}10^{-8}}
\newcommand{\friedmanCD}{3.0}
\newcommand{\friedmanRankERM}{5.78}
\newcommand{\friedmanRankMCD}{6.11}
\newcommand{\friedmanRankSWA}{4.94}
\newcommand{\friedmanRankDeepEns}{5.17}
\newcommand{\friedmanRankBagTwo}{2.89}
\newcommand{\friedmanRankBagFive}{1.67}
\newcommand{\friedmanRankTwin}{1.44}
\newcommand{\friedmanRankGapBagFiveTwin}{0.22}
\newcommand{\triageKtenLow}{58}
\newcommand{\triageKtenHigh}{100}
\newcommand{\triageKtwoLow}{48}
\newcommand{\triageKtwoHigh}{83}
\newcommand{\triageKtwoToKtenGapLow}{10}
\newcommand{\triageKtwoToKtenGapHigh}{24}
\newcommand{\fragTopLow}{25}
\newcommand{\fragTopHigh}{60}
\newcommand{\entropyTopLow}{21}
\newcommand{\entropyTopHigh}{47}
\newcommand{\entropyGapMin}{1}
\newcommand{\entropyGapMax}{13}
\newcommand{\bayesTwinTrials}{50}
\newcommand{\bayesTwinInitTrials}{4}
\newcommand{\bayesTwinKernel}{\textrm{Mat\'ern-2.5}}
\newcommand{\bayesTwinLamMin}{10^{-3}}
\newcommand{\bayesTwinLamMax}{10^{4}}
\newcommand{\churnPenalty}{100}
\newcommand{\bayesTwinBeatsRuleCount}{4}
\newcommand{\bayesTwinTiesRuleCount}{2}
\newcommand{\bayesTwinLosesRuleCount}{3}
\newcommand{\bayesTwinTiesOrLosesCount}{5}
\newcommand{\bayesTwinMatchesOrBeatsCount}{6}
\newcommand{\boTopkK}{10}
\newcommand{\boTopkErmJaccardLow}{0.03}
\newcommand{\boTopkErmJaccardHigh}{0.56}
\newcommand{\boTopkTwinJaccardLow}{0.45}
\newcommand{\boTopkTwinJaccardHigh}{0.69}
\newcommand{\boTopkTwinDeltaMin}{+0.12}
\newcommand{\boTopkTwinDeltaMax}{+0.56}

\newcommand{\boLoopRegK}{20}
\newcommand{\boLoopRegInitN}{50}
\newcommand{\boLoopRegBudget}{10}
\newcommand{\boLoopRegLam}{3}
\newcommand{\boLoopRegErmStdLow}{0.082}
\newcommand{\boLoopRegErmStdHigh}{0.547}
\newcommand{\boLoopRegErmJaccardLow}{0.37}
\newcommand{\boLoopRegErmJaccardHigh}{0.48}
\newcommand{\boLoopRegErmStdRangePctLow}{1.27}
\newcommand{\boLoopRegErmStdRangePctHigh}{2.53}
\newcommand{\boLoopRegBagStdLow}{0.000}
\newcommand{\boLoopRegBagStdHigh}{0.178}
\newcommand{\boLoopRegBagJaccardLow}{0.45}
\newcommand{\boLoopRegBagJaccardHigh}{0.65}

\newcommand{\boLoopRegTwinStdLow}{0.000}
\newcommand{\boLoopRegTwinStdHigh}{0.362}
\newcommand{\boLoopRegTwinJaccardLow}{0.39}
\newcommand{\boLoopRegTwinJaccardHigh}{0.61}

\newcommand{\boLoopRegEsolErmFinalLo}{1.10}
\newcommand{\boLoopRegEsolErmFinalHi}{1.57}

\newcommand{\boLoopRegNumDatasets}{3}

\newcommand{\boLoopRegBagStdRedWins}{2}
\newcommand{\boLoopRegTwinStdRedLow}{34}
\newcommand{\boLoopRegTwinStdRedHigh}{100}

\newcommand{\regTwinLow}{39}
\newcommand{\regTwinHigh}{42}
\newcommand{\regBagTwoLow}{29}
\newcommand{\regBagTwoHigh}{32}
\newcommand{\regBagFiveLow}{55}
\newcommand{\regBagFiveHigh}{58}
\newcommand{\regFragRatioLow}{42}
\newcommand{\regFragRatioHigh}{57}
\newcommand{\chembertaAccDropLow}{11}
\newcommand{\chembertaAccDropHigh}{18}
\newcommand{\chembertaCutLow}{9}
\newcommand{\chembertaCutHigh}{76}
\newcommand{\chembertaLamThreeHundredFailCount}{5}
\newcommand{\chembertaLamThreeHundredDenom}{6}
\newcommand{\waterbirdsAccCollapse}{26}
\newcommand{\waterbirdsLamTenCut}{55}
\newcommand{\ginBagCutLamThreeHundred}{54}
\newcommand{\ginBagAccGain}{+4.5}
\newcommand{\ginAccDropLamThreeHundred}{16}
\newcommand{\ginCutLamTen}{52}
\newcommand{\ginRuleLamCutPct}{51.8}
\newcommand{\ginRuleLam}{10}
\newcommand{\ginRuleLamDeltaLo}{-13.2}
\newcommand{\ginRuleLamDeltaHi}{-10.6}
\newcommand{\ginRuleLamAccGain}{+0.5}
\newcommand{\borderlineGapMin}{3}
\newcommand{\borderlineGapMax}{4}
\newcommand{\borderlineNTestMin}{57}
\newcommand{\borderlineNTestMax}{104}
\newcommand{\nScalingSlope}{-0.20}
\newcommand{\nScalingSymKLLow}{0.83}
\newcommand{\nScalingSymKLMin}{0.62}
\newcommand{\nScalingSymKLEnd}{0.69}
\newcommand{\nScalingMmin}{800}

\newcommand{\diliNTest}{76}

\newcommand{\cypTwoOverallChurn}{13.3}
\newcommand{\cypTwoPosFrac}{0.30}
\newcommand{\diliOverallChurn}{16.8}
\newcommand{\imbalPosFracHigh}{0.78}
\newcommand{\ginErmAcc}{0.742}
\newcommand{\ginErmAccTolerance}{0.722}
\newcommand{\ginSymKLReductionPct}{99}
\newcommand{\mofThermalConvLow}{48}
\newcommand{\mofThermalConvHigh}{58}
\newcommand{\regIdMaeImproveLow}{3}
\newcommand{\regIdMaeImproveHigh}{9}

\newcommand{\twinWinsBagTwoHeldout}{7}
\newcommand{\twinWinsBagTwoHeldoutDenom}{8}
\newcommand{\twinWinsBagFiveHeldout}{5}
\newcommand{\twinWinsBagFiveHeldoutDenom}{8}
\newcommand{\bagFiveWinsTwinHeldout}{3}
\newcommand{\maxAcrossSeedChurnSpreadPP}{5.6}
\newcommand{\maxAcrossSeedChurnSpreadCell}{BACE, ERM}
\newcommand{\maxAcrossSeedDeltaSpreadPP}{3.9}
\newcommand{\maxAcrossSeedDeltaSpreadCell}{BACE, twin-bootstrap $\lambda{=}300$}

%% file: sections/abstract.tex
\begin{abstract}
Scientific machine learning reports predictive performance.  It does not report whether the same
prediction would survive a different draw of training data.  Across $\nDatasetsHeadline$ chemistry benchmarks,
two classifiers trained on independent bootstraps of the same
training set agree on aggregate accuracy to within
$\accDiffMin\text{--}\accDiffMax$ percentage points but disagree on
the class label of $\churnMin\text{--}\churnMax\%$ of test
molecules.  We call this gap \emph{cross-sample prediction churn}.
The standard parameter-side techniques (deep ensembles, MC dropout, stochastic weight averaging)
do not reduce this gap; two data-side methods do.  The first is
$K$-bootstrap bagging, which cuts the rate
$\bagFiveLow\text{--}\bagFiveHigh\%$ on every dataset at no accuracy
cost ($K{\times}$-ERM compute).  The second is \emph{twin-bootstrap},
our proposal: two networks trained jointly on independent bootstraps
with a sym-KL consistency loss between their predictions, which at
matched $2{\times}$-ERM compute reduces churn a further median
$\twinFurtherMedianReductionVsBagTwo\%$ beyond bagging-$K{=}2$.
Cross-sample prediction churn deserves a column alongside
predictive performance in scientific-ML benchmark reports, because
without it the parameter-side and data-side methods are
indistinguishable on the metric they actually differ on.
\end{abstract}

%% file: sections/introduction.tex
\section{Introduction}
\label{sec:introduction}

\begin{figure}[t]
  \centering
  \includegraphics[width=\linewidth]{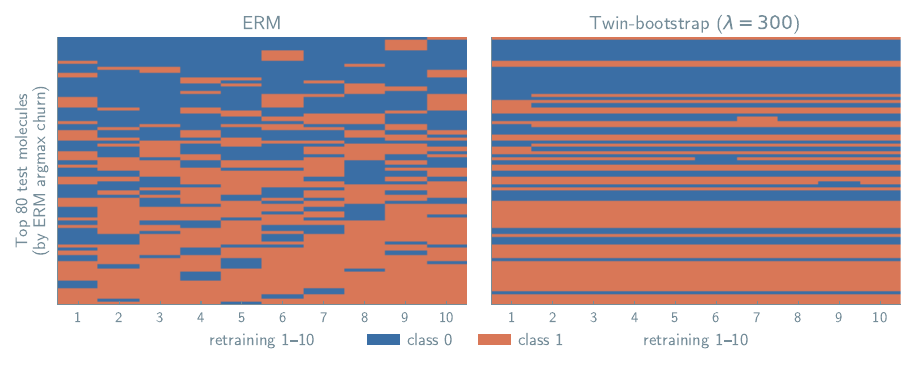}
  \caption{\textbf{Twin-bootstrap eliminates most of ERM's
    retraining-induced prediction flips on BACE.}  Each row is one
    of the $80$ test molecules with the largest cross-sample
    contrast; each column is one of ten retrainings on an
    independent bootstrap of the BACE training pool; cells
    are coloured by predicted class.  Visible vertical stripes
    in the left (ERM) panel are predictions that flip class across
    retrainings; under twin-bootstrap (right; $\lambda{=}300$, matched
    compute against bagging-$K{=}2$) the same molecules become
    near-uniform.  The mean class-flip rate over the full test set drops
    from $\baceErmChurn\%$ (ERM) to $\baceTwinChurn\%$ (twin-bootstrap).}
  \label{fig:overview}
\end{figure}

A bioactivity classifier is retrained on a new batch of assay results.
Aggregate accuracy moves by $\baceMeanAccuracyDiff$
percentage points; the class label flips on
$\baceErmChurn\%$ of test molecules
(Figure~\ref{fig:overview}).  We call this \emph{cross-sample
prediction churn} --- the fraction of test predictions that change
class between two models trained on independent samples of the same
training population.
This is a per-prediction stability gap that aggregate-performance reporting in scientific-ML benchmarks systematically
hides.  In closed-loop laboratories, Bayesian-optimisation campaigns,
and virtual-screening pipelines, model predictions feed directly into
experimental decisions: the molecule selected for synthesis, the
candidate prioritised in a screening campaign, the hit returned by a
virtual screen.  Churn is
the rate at which retraining changes which molecules the model
selects, and is therefore the operational stability the practitioner
cares about.

Chemistry models are designed to be invariant to atomic permutation,
rotation, and translation; $E(3)$-equivariant message passing and
permutation-equivariant aggregators are standard.  \emph{Sample invariance} --- invariance to which independent
draw of the training population the model was fit on --- is the
analogue on the data axis, but is not routinely measured in scientific
ML.  However, a practitioner who updates a screening set, retrains,
and reroutes synthesis based on the new top predictions confronts the
challenge of cross-sample churn on every dataset update.

Two existing approaches are conceptually nearby but do not measure
or reduce this gap.  \emph{Prediction churn}~\citep{launch_and_iterate,
bhojanapalli2021reproducibility, jiang2022} measures
retraining-induced disagreement on web-scale data; prior work varies
training conditions on a fixed dataset or compares a candidate model
against a deployed incumbent.  In scientific ML there is often no
incumbent, and the dominant source of variance is the small training
set itself, redrawn between retrainings.  \emph{Epistemic uncertainty} tools --- deep
ensembles~\citep{lakshminarayanan2017simple}, Monte Carlo (MC)
dropout~\citep{gal2015dropout}, and stochastic weight averaging
(SWA)~\citep{izmailov2018averaging} --- sample over weights at fixed
data and so capture the parameter-side slice of model variance, but
not the data-sampling slice we measure: across our
$\nDatasetsHeadline$ chemistry benchmarks, the three together shift
the class-flip rate by $\paramSideLow\%$ to $\paramSideHigh\%$
relative to empirical risk minimisation (ERM), with no consistent
sign.  Methods that vary the data --- bagging~\citep{Breiman1996} and
the twin-bootstrap procedure we propose --- do reduce it.

\paragraph{Contributions.}
\begin{itemize}\itemsep0pt
  \item \textbf{Cross-sample prediction churn.}  The data-sampling
        analogue of model variance --- the slice of uncertainty that
        deep ensembles, MC dropout, and SWA all miss because they
        sample over weights at fixed data
        (Section~\ref{sec:measurement}).
  \item \textbf{Magnitudes.}  On \nDatasetsHeadline{} chemistry
        benchmarks (MoleculeNet, TDC ADME and Tox, materials-science),
        per-prediction churn flips
        $\churnMin\text{--}\churnMax\%$ of test predictions while
        aggregate accuracy moves only
        $\accDiffMin\text{--}\accDiffMax$\,pp
        (Table~\ref{tab:fragility-magnitudes}).
  \item \textbf{Bagging.}  $K$-bootstrap bagging~\citep{Breiman1996}
        cuts churn $\bagFiveLow\text{--}\bagFiveHigh\%$ on every
        dataset at no accuracy cost, at $K\times$-ERM compute
        (Table~\ref{tab:main}).
  \item \textbf{Twin-bootstrap.}  Two networks are trained jointly
        on independent bootstraps with a symmetric-KL consistency
        loss between their predictions ($\sim\overlapPctMid\%$
        inter-network data overlap by construction).  At matched
        $2\times$-ERM compute it cuts churn a further median
        $\twinFurtherMedianReductionVsBagTwo\%$ beyond bagging-$K{=}2$
        and matches $5\times$-compute bagging-$K{=}5$ in mean rank
        (Table~\ref{tab:main}, Figure~\ref{fig:main}); the same
        reduction shows up at the Bayesian-optimisation acquisition
        layer (Appendices~\ref{app:bo_topk} and~\ref{app:bo_loop_reg}).
  \item \textbf{Per-prediction triage.}  On top of any deployed
        model, sorting by per-example churn from a single extra
        retraining captures
        $\triageKtenLow\text{--}\triageKtenHigh\%$ of class flips at
        the top-$\reviewFraction\%$ review fraction; better than
        predictive entropy on every dataset
        (Section~\ref{sec:churn-decile}).
\end{itemize}

%% file: sections/related_work.tex
\section{Related work}
\label{sec:related}

\paragraph{Prediction churn.}
\citet{launch_and_iterate} introduced \emph{prediction churn} as the
disagreement between two classifiers trained on the same task with
different runs or data updates; their MCMC stabiliser and
\citet{jiang2022}'s distillation-from-incumbent are deployment-loop
methods that compare a candidate model against a frozen incumbent on
web-scale data.  \citet{bhojanapalli2021reproducibility} instead
hold the training data fixed and isolate parameter-side variance
(initialisation, mini-batch order, hardware) as the source of
churn on CIFAR/ImageNet.

\paragraph{Codistillation operating points on the data-overlap axis.}
Two-network sym-KL consistency has been applied at two operating
points on the inter-network data-overlap axis:
\citet{anil2018large} train on disjoint shards ($0\%$ overlap) with
a KL agreement loss for distributed-training throughput;
\citet{bhojanapalli2021reproducibility} share the entire training
set ($100\%$ overlap) for parameter-side reproducibility.
Our twin-bootstrap inherits the sym-KL loss form and applies it at a
third operating point: independent bootstraps of the canonical
training pool, sharing ${\sim}40\%$ of indices in expectation
(Appendix~\ref{app:overlap_spectrum}).

\paragraph{Predictive multiplicity and the Rashomon set.}
The Rashomon set~\citep{Breiman2001} is the set of near-loss-equivalent
models on a fixed dataset; \citet{marx2020} operationalise this as
\emph{predictive multiplicity}, \citet{damour2022} attribute it to
under-specified pipelines, and \citet{black2022} build selective
ensembles that abstain when ensemble consistency cannot be certified
by a hypothesis test.  These works hold the dataset fixed and vary
models within the loss-equivalent class --- orthogonal to
cross-sample churn we discuss here, where the dataset varies and each retraining
is one point in the resulting Rashomon set.

\paragraph{Parameter-side uncertainty and bagging.}
Deep ensembles~\citep{lakshminarayanan2017simple} and Monte Carlo
dropout~\citep{gal2015dropout} sample over model parameters at fixed
data.  Bagging~\citep{Breiman1996} samples over the data --- an axis
that has been studied for accuracy in deep learning but, to our knowledge, not for
cross-sample stability.

\paragraph{Clinical-biostatistics precedent.}
\citet{Riley2023} define four levels of stability in clinical
prediction-model risk estimates (mean, distribution, subgroups,
individuals) and assess the individual level by retraining on
$1000$ bootstrap samples of the development data, reporting
prediction-instability plots and a mean absolute prediction error.
They demonstrate considerable instability of individualised risk
estimates and argue that stability checks should be a routine part
of model development --- the closest published precedent for the
argument we make in Section~\ref{sec:discussion}, in a different
domain and without a deep-learning treatment.

%% file: sections/measurement.tex
\section{Cross-sample prediction churn}
\label{sec:measurement}

\paragraph{Setting.}
A learner $\mathcal{A}$ maps a training set
$S = \{(x_i, y_i)\}_{i=1}^{N}$ drawn iid from population $\mathcal{D}$
to a classifier $f_S = \mathcal{A}(S)$ whose output $f_S(x) \in
\mathbb{R}^{|\mathcal{C}|}$ scores the classes $c \in \mathcal{C}$;
the predicted label at $x$ is
$\hat{y}(x) = \arg\max_{c} f_S(x)_c$.
Cross-sample churn quantifies how $\hat{y}(x)$ changes when $S$ is
replaced by an independent iid draw from $\mathcal{D}$.

\paragraph{Definition.}
Given two iid training samples $S_A, S_B \sim \mathcal{D}^N$ and a
test example $x$, the \emph{cross-sample churn} of $\mathcal{A}$ at
$x$ is
\begin{equation}
  \rho(\mathcal{A}, x) =
  \mathbb{P}_{S_A, S_B}\!\big[\arg\max f_{S_A}(x) \neq \arg\max f_{S_B}(x)\big].
\end{equation}
A churn-aware report supplements the standard predictive-performance
columns with the expected per-example argmax disagreement
$\bar\rho(\mathcal{A}) = \mathbb{E}_x [\rho(\mathcal{A}, x)]$, and
optionally with a distributional analogue
$\bar\rho_{\mathrm{KL}}(\mathcal{A})
= \mathbb{E}_x \big[\tfrac{1}{2}(\mathrm{KL}(f_{S_A}(x) \| f_{S_B}(x))
                                + \mathrm{KL}(f_{S_B}(x) \| f_{S_A}(x)))\big]$.
We use ``cross-sample churn'' (or just ``churn'' when context is
clear) for $\bar\rho$ and ``class-flip rate'' for the same quantity
reported as a percentage; the two names refer to the same quantity.

\paragraph{Bootstrap estimator.}
A practitioner has one observed training set $S$, not a population
$\mathcal{D}$.  We estimate $\rho$ by drawing two independent
bootstraps from $S$ (size $N$, with replacement); each bootstrap is
a draw from the empirical distribution $\hat{\mathcal{D}}_S$, which
converges to $\mathcal{D}$ as $N \to \infty$.  All
numerical magnitudes we report are with respect to this bootstrap
proxy, which is computable from a single observed training
set.

\paragraph{Measurement protocol.}
A \emph{canonical seed} fixes one train/test split.  Within a
canonical seed, the training pool and test set are reused across
every method and train seed, so test-side variance is zero by
construction.  For each train seed $s \in \{1, \dots, \nSeeds\}$ we
draw a bootstrap $S^{(s)}$ from the canonical training pool and
train $f^{(s)} = \mathcal{A}(S^{(s)})$ on it.  Cross-sample churn is the average of
$\mathbf{1}[\arg\max f^{(s)}(x) \neq \arg\max f^{(s')}(x)]$ over all
$\binom{\nSeeds}{2}{=}\nSeedPairs$ pairs $(s, s')$ and over the canonical test set.
Confidence intervals come from $10{,}000$-sample bootstrap on the
seed-pair distribution; cross-method deltas are paired.

For the chemistry-MLP comparison we run $\nCanonicalSeeds$
replicates; quoted magnitudes are across-replicate means and the $\lambda$-selection
rule picks the same value on every replicate
(Appendix~\ref{app:seed_sensitivity}).

\paragraph{Datasets.}
\label{sec:datasets}
We evaluate on seventeen chemistry benchmarks
($N \in [\,304, 4658\,]$): MoleculeNet~\citep{Wu2018}
(BACE, BBBP, ClinTox); the TDC~\citep{huang2021therapeutics} ADME
and Tox suites (HIA\_Hou, Bioavailability\_Ma, Pgp\_Broccatelli,
BBB\_Martins, CYP\{2C9,2D6,3A4\}-Sub, hERG, DILI, AMES,
Skin\_Reaction); and three materials-science
benchmarks (TADF~\citep{jablonka2026clever, Huang2024},
MOF-thermal-stability, and MOF-solvent-removal~\citep{jablonka2026clever, Nandy2022}).
We require ERM accuracy to exceed the majority-class baseline by at
least $\preregFilterPP$\,pp on a canonical test set of at least
$\preregMinTestSize$ examples (threshold fixed before any
cross-sample analysis); this keeps nine datasets for the method
comparison, flags three borderline datasets, and excludes five
(per-dataset outcomes in Appendix~\ref{app:filter_outcomes}).  
Featurisation follows standard chemistry conventions for each dataset: $2048$-bit Morgan radius-2
fingerprints for MoleculeNet/TDC; $+217$ RDKit descriptors for
TADF; $174$ RAC descriptors plus geometric solvent-accessibility
features for MOFs~\citep{Moosavi2020}.  The default architecture is
a $256$-unit MLP trained from scratch on these features.  For each
MoleculeNet/TDC dataset we group molecules by Bemis--Murcko scaffold
and partition the within-scaffold-group pool $80/20$ into train and
id-test under the canonical seed; TADF and the MOF benchmarks use a
random $80/20$ split.

%% file: sections/magnitudes.tex
\section{\boldmath Cross-sample churn flips $\churnMin$--$\churnMax\%$ of test predictions on \nDatasetsHeadline{} chemistry benchmarks}
\label{sec:churn-magnitudes}

Table~\ref{tab:fragility-magnitudes} reports the cross-bootstrap
class-flip rate and the corresponding distributional gap (sym-KL)
for each of the \nDatasetsHeadline{} chemistry datasets that pass the
$+\preregFilterPP$pp ERM-vs-majority filter -- on datasets where ERM
fails to beat majority-class prediction by at least
$\preregFilterPP$pp, cross-sample churn conflates the method shifting
its decision boundary with the majority class itself shuffling under
sampling noise.  Excluded datasets are listed in
Appendix~\ref{app:filter_outcomes}.  Alongside the churn columns we
report training-set size $N$, ERM accuracy, and the mean pairwise
accuracy difference $|\Delta\text{acc}|$ between two retrainings.  Class-flip rate ranges from $\churnMin\%$ (BBB-Martins)
to $\churnMax\%$ (MOF-thermal).  Aggregate accuracy moves
$\accDiffMin\text{--}\accDiffMax$\,pp between two retrainings while
individual predictions disagree at
$\churnAccRatioMin\text{--}\churnAccRatioMax\times$ that rate.
Sym-KL varies by ${\sim}\symKLDatasetSpread\times$ across datasets,
and the variation is task-dependent rather than a function of
training-set size alone (e.g.\ TADF and BACE both have
$N \approx 1000$ but differ by ${\sim}2\times$ in sym-KL).
At these magnitudes, $\churnMin\text{--}\churnMax\%$ of the
per-molecule decisions a downstream pipeline would route to
synthesis or screening change between two retrainings on the same
training population --- a reshuffling that is invisible in
aggregate metrics such as mean accuracy.  Two acquisition-layer ablations confirm this carries through to
Bayesian optimisation: the top-$\boTopkK$ Jaccard overlap of ERM's
acquisition shortlist across retrainings sits at
$\boTopkErmJaccardLow$--$\boTopkErmJaccardHigh$ on chemistry
benchmarks (Appendix~\ref{app:bo_topk}), and on regression the
cross-trajectory standard deviation of the BO final-best $y$ under
ERM reaches $\boLoopRegErmStdRangePctLow$--$\boLoopRegErmStdRangePctHigh\%$
of the response range (Appendix~\ref{app:bo_loop_reg}).

\input{sections/tables/fragility_magnitudes.tex}

The within-task $N$-scaling on BACE (Appendix~\ref{app:nscaling})
shows sym-KL trending downward with a log-log slope of
$\nScalingSlope$ in $M$; argmax churn is noisier and does not
decrease monotonically with $M$.

%% file: sections/tables/fragility_magnitudes.tex
\begin{table}[t]
  \centering
  \caption{\textbf{\boldmath Two retrainings on independent bootstraps differ in aggregate accuracy by 1.3\,--\,4.2\,pp on average, but disagree on $8\text{--}22\%$ of individual test predictions.}  Cross-bootstrap class-flip rate on the canonical id-test of the nine chemistry datasets that pass a $+5$pp ERM-vs-majority filter on test sets of at least $60$ examples (BACE is the development dataset; the other eight are held-out).  Three datasets that pass the filter only marginally are reported in Appendix~\ref{app:borderline}.  ERM id-acc is the mean across $10$ retrainings; $|\Delta\text{acc}|$ is the mean absolute accuracy difference between two retrainings, averaged over the same $45$ pairs of $10$ retrainings as the churn column.  Class-flip rate is the per-example argmax-disagreement rate (cross-sample churn); sym-KL is the corresponding distributional gap.  All paired columns report mean with $95\%$ paired-bootstrap CIs ($10{,}000$ resamples).}
  \label{tab:fragility-magnitudes}
  \scriptsize
  \resizebox{\linewidth}{!}{%
  \begin{tabular}{lrrrl@{\hspace{1em}}ll}
    \toprule
    & & & \multicolumn{2}{c}{Aggregate accuracy} & \multicolumn{2}{c}{Per-prediction disagreement} \\
    \cmidrule(lr){4-5} \cmidrule(lr){6-7}
    Dataset & $N_{\text{train}}$ & $N_{\text{id-test}}$ & ERM id-acc & $|\Delta\text{acc}|$ (pp) & Argmax churn (\%) & Sym-KL (nats) \\
    \midrule
    DILI & 304 & 76 & 0.724 [0.671, 0.776] & 4.1 [3.4, 4.8] & \textbf{16.8 [15.8, 17.7]} & \textbf{0.751 [0.692, 0.807]} \\
    CYP2D6-Sub & 427 & 106 & 0.752 [0.698, 0.811] & 4.2 [3.4, 4.9] & \textbf{13.3 [12.6, 14.1]} & \textbf{0.640 [0.582, 0.701]} \\
    Pgp & 780 & 194 & 0.842 [0.820, 0.871] & 2.0 [1.6, 2.4] & \textbf{10.3 [9.7, 10.9]} & \textbf{0.512 [0.478, 0.547]} \\
    BACE\,(dev) & 968 & 242 & 0.779 [0.756, 0.810] & 1.8 [1.5, 2.2] & \textbf{16.1 [15.6, 16.7]} & \textbf{0.753 [0.712, 0.796]} \\
    TADF & 1007 & 428 & 0.806 [0.787, 0.822] & 1.3 [1.1, 1.5] & \textbf{12.7 [12.3, 13.1]} & \textbf{0.401 [0.384, 0.418]} \\
    MOF-thermal & 1251 & 627 & 0.718 [0.703, 0.745] & 1.5 [1.2, 1.9] & \textbf{21.8 [21.2, 22.4]} & \textbf{0.383 [0.368, 0.398]} \\
    BBB-Martins & 1300 & 324 & 0.867 [0.849, 0.895] & 1.7 [1.4, 2.1] & \textbf{8.0 [7.6, 8.4]} & \textbf{0.500 [0.477, 0.524]} \\
    BBBP & 1305 & 326 & 0.838 [0.819, 0.868] & 1.5 [1.1, 1.8] & \textbf{8.5 [8.1, 8.9]} & \textbf{0.470 [0.444, 0.496]} \\
    AMES & 4658 & 1164 & 0.779 [0.756, 0.799] & 1.6 [1.3, 1.9] & \textbf{15.2 [14.8, 15.5]} & \textbf{1.121 [1.092, 1.153]} \\
    \bottomrule
  \end{tabular}
  }
\end{table}

%% file: sections/methods.tex
\section{Methods}
\label{sec:methods}

We compare six approaches under the protocol of
Section~\ref{sec:measurement}.  All methods share the same
architecture and optimiser unless stated otherwise: a $256$-unit,
two-hidden-layer MLP with ReLU activations on the dataset-specific
input features (Section~\ref{sec:measurement}), trained from
scratch with AdamW~\citep{loshchilov2019decoupled} (weight decay
$10^{-4}$, gradient clipping at $1.0$), learning rate $10^{-3}$,
batch size $64$, and $30$ epochs.

\paragraph{ERM.}
A single model trained with AdamW on one bootstrap of the canonical
training set.

\paragraph{Stochastic weight averaging (SWA)~\citep{izmailov2018averaging}.}
Trained as ERM, but at the end of every epoch in the second half of
training we snapshot model weights and accumulate a running average.
The averaged weights replace the final-epoch weights at inference time;
prediction is a single forward pass through the averaged-weight
model.  SWA is a parameter-side smoother: it modifies the weight
trajectory without varying the data sample.  Cross-sample churn is
measured between two such weight-averaged models trained on
independent bootstraps.

\paragraph{MC dropout.}
A single model with dropout layers ($p=0.2$) inserted after each
hidden ReLU, trained by SGD on one bootstrap.  At inference time, $T=20$
stochastic forward passes are averaged with dropout active
\citep{gal2015dropout}.  This captures dropout-mask variance at fixed
data; cross-sample churn is then measured between the
$T$-averaged predictions of two models trained on independent
bootstraps.

\paragraph{Deep ensemble.}
$K$ models trained on the \emph{same} bootstrap with different
initialisation seeds; predictions are averaged at inference time.  This
captures parameter-distribution variance only.

\paragraph{\boldmath$K$-bootstrap bagging.}
The classical estimator of \citet{Breiman1996}: $K$ models
trained on $K$ independent bootstraps of the canonical training set,
predictions averaged at inference time.  No auxiliary loss; the data
axis is varied along with parameters.

\paragraph{Twin-bootstrap.}
Two networks $\theta_A, \theta_B$ are trained jointly on independent
bootstraps $S_A, S_B$.  At each step we draw one mini-batch from each
loader, forward both networks on both mini-batches, and minimise
\begin{equation}
  \mathcal{L} \;=\;
    \mathcal{L}_{\mathrm{CE}}(\theta_A; S_A^{(t)}) \;+\;
    \mathcal{L}_{\mathrm{CE}}(\theta_B; S_B^{(t)}) \;+\;
    \lambda \cdot \mathcal{L}_{\mathrm{cons}},
\end{equation}
where $\mathcal{L}_{\mathrm{CE}}$ is standard cross-entropy on each
network's own bootstrap, and the consistency term
$\mathcal{L}_{\mathrm{cons}}$ is the symmetric KL between the two
networks' softmax distributions evaluated on the union of the two
mini-batches.  At inference time, predictions are averaged.

\paragraph{Mechanism.}
At $\lambda{=}0$ the procedure is $K{=}2$ bagging.  At $\lambda > 0$
the consistency term penalises divergence on every example seen in
either mini-batch, including those that appear in both bootstraps.
Bootstrap-with-replacement at size $N$ covers ${\sim}63\%$ of unique
indices in expectation, so two independent bootstraps share
${\sim}0.63^2 \approx 40\%$ of indices.  The consistency loss
therefore acts on the overlap while the cross-entropy losses
specialise on the non-shared remainder.  
Algorithm~\ref{alg:twin-bootstrap} gives the per-epoch
procedure.

\begin{algorithm}[H]
  \caption{Twin-bootstrap training (one epoch).}
  \label{alg:twin-bootstrap}
  \begin{algorithmic}[1]
    \Require canonical training set $S$, networks $\theta_A, \theta_B$,
             optimisers $\mathrm{opt}_A, \mathrm{opt}_B$, weight $\lambda$
    \State Sample $S_A, S_B \sim \mathrm{Bootstrap}(S)$
           \Comment{$N$ draws with replacement, independent}
    \State Build dataloaders $L_A$ over $S_A$, $L_B$ over $S_B$
    \For{$(B_A, B_B)$ zipped from $(L_A, L_B)$}
      \State $\hat{p}_A^{(A)} \gets f_{\theta_A}(B_A);\quad \hat{p}_B^{(A)} \gets f_{\theta_B}(B_A)$
      \State $\hat{p}_A^{(B)} \gets f_{\theta_A}(B_B);\quad \hat{p}_B^{(B)} \gets f_{\theta_B}(B_B)$
      \State $\mathcal{L}_{\mathrm{CE}} \gets \mathrm{CE}(\hat{p}_A^{(A)}, y_A) + \mathrm{CE}(\hat{p}_B^{(B)}, y_B)$
      \State $\mathcal{L}_{\mathrm{cons}} \gets \tfrac{1}{2}\big[\mathrm{symKL}(\hat{p}_A^{(A)}, \hat{p}_B^{(A)}) + \mathrm{symKL}(\hat{p}_A^{(B)}, \hat{p}_B^{(B)})\big]$
      \State $\mathcal{L} \gets \mathcal{L}_{\mathrm{CE}} + \lambda \, \mathcal{L}_{\mathrm{cons}}$
      \State Backprop $\mathcal{L}$ through both networks; step $\mathrm{opt}_A, \mathrm{opt}_B$
    \EndFor
  \end{algorithmic}
\end{algorithm}

\paragraph{Hyperparameter selection.}
A single development dataset (BACE) is used to select $\lambda$ via
the pre-registered rule: largest $\lambda$ in
$\{1, 3, 10, 30, 100, 300\}$ such that BACE id-accuracy is within
$\preregTolerance$ of ERM id-accuracy.  The rule yields $\lambda{=}300$
(Pareto curve in Appendix~\ref{app:pareto}).
We apply $\lambda{=}300$ unchanged to every held-out dataset; per-dataset
Bayesian optimisation of $\lambda$ with a cross-sample-churn objective
matches or beats this frozen choice on
$\bayesTwinMatchesOrBeatsCount/\nDatasetsHeadline$ datasets and
strictly improves on
$\bayesTwinBeatsRuleCount/\nDatasetsHeadline$
(Appendix~\ref{app:bayes_twin}).
$K$ for ensemble methods is set a priori to $5$ following
\citet{lakshminarayanan2017simple}; we additionally report $K{=}2$ for
compute-matched comparison against twin-bootstrap.

%% file: sections/experiments.tex
\section{Empirical evaluation}
\label{sec:experiments}

\begin{figure}[t]
  \centering
  \includegraphics[width=\linewidth]{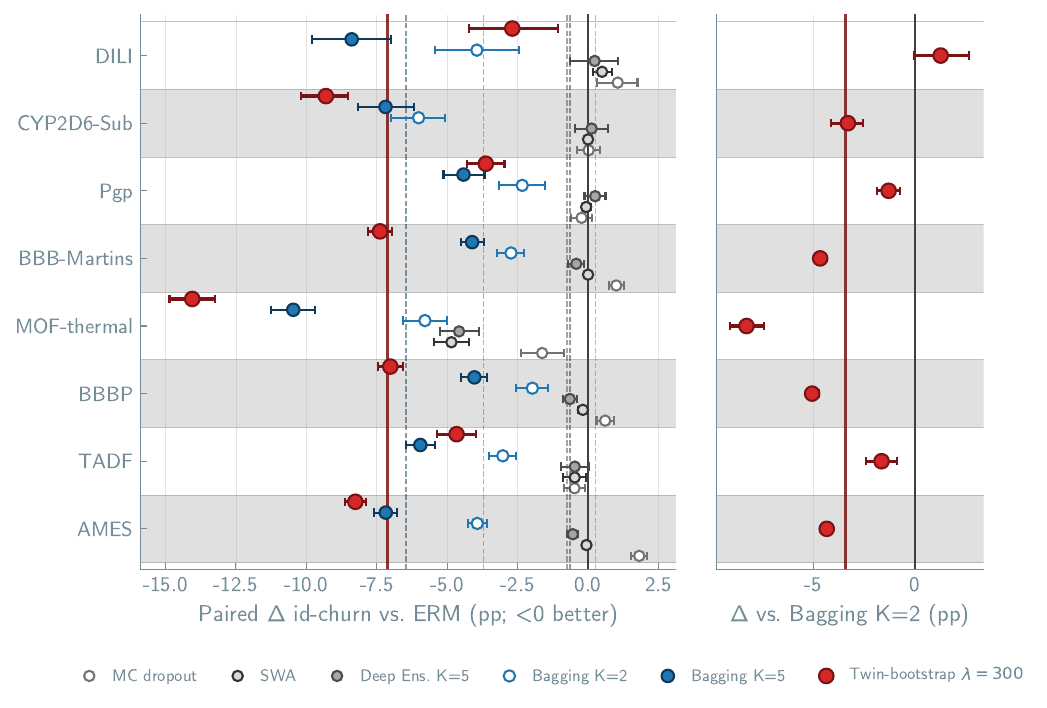}
  \caption{\textbf{Bagging and twin-bootstrap beat ERM on every chemistry
    benchmark; MC dropout, deep ensembles, and SWA do not.}  Left: paired
    $\Delta$ id-churn vs.\ ERM for six methods, one row per dataset
    (smallest $N$ at top), $95\%$ paired-bootstrap CIs across
    the $\nSeedPairs$ seed pairs.  Vertical reference lines mark
    each method's across-dataset mean (twin-bootstrap solid, others
    dashed; colours match markers); the solid black line is parity
    with ERM.  Right: $\Delta$ vs.\ matched-compute bagging-$K{=}2$,
    twin-bootstrap only; CI excludes zero on
    $\twinWinsBagTwoHeldout/\twinWinsBagTwoHeldoutDenom$ datasets and
    ties on DILI.  Frozen $\lambda{=}300$ is selected on the BACE development
    dataset alone; no held-out tuning.}
  \label{fig:main}
\end{figure}

\paragraph{Cross-sample reduction across benchmarks.}
Across the \nDatasetsHeadline{} chemistry benchmarks (BACE plus
\nDatasetsHeldout{} held-out), bagging-$K{=}5$ and twin-bootstrap both
beat ERM on the cross-sample class-flip rate with $95\%$ paired-bootstrap CIs
that exclude zero (Table~\ref{tab:main}, Figure~\ref{fig:main}).  The
three parameter-side techniques --- MC dropout, deep ensembles, and
stochastic weight averaging (SWA) --- do not reduce churn
consistently.  Their paired $\Delta$ class-flip rate vs.\ ERM
ranges $\paramSideLow\%$ to $\paramSideHigh\%$ across datasets,
with no consistent sign.  Averaging $T{=}20$ stochastic forward passes
(MC dropout), averaging $K{=}5$ networks at fixed data
(deep ensembles), or averaging weight snapshots from a single
training trajectory (SWA) all keep the data-resampling axis
constant.  
At matched compute against bagging-$K{=}2$, twin-bootstrap wins on
$\twinWinsBagTwoHeldout/\twinWinsBagTwoHeldoutDenom$ held-out
datasets (paired CI excludes zero); the comparison ties on DILI,
whose canonical id-test holds only $\diliNTest$ examples.  Against
the $5\times$-compute bagging-$K{=}5$, twin-bootstrap wins on
$\twinWinsBagFiveHeldout/\twinWinsBagFiveHeldoutDenom$ held-out
datasets and trails on $\bagFiveWinsTwinHeldout$ (DILI, Pgp, TADF).
The aggregate Friedman test rejects equal-ranks at
$\chi^2 = \friedmanChi$, $p = \friedmanP$; twin-bootstrap ranks
first, indistinguishable from bagging-$K{=}5$ at less than half the
compute (Appendix~\ref{app:friedman}).  What reduces churn turns out
to be different from what improves accuracy: deep ensembles average
five independently initialised networks at fixed data and raise mean
accuracy by $\deepEnsAccDelta$\,pp without changing churn, while
twin-bootstrap varies the data sample and cuts churn a median
$\twinMedianReduction\%$ at a small accuracy cost
($\twinAccDelta$\,pp on average).

\input{sections/tables/main.tex}  

\paragraph{Distributional disagreement.}
Argmax churn counts only whether the top-predicted class changes; it
ignores how the rest of the probability distribution shifts.  Symmetric
KL between two bootstraps captures that and distinguishes methods more
finely: bagging-$K{=}5$ reduces it
$\symKLBagFoldLow\text{--}\symKLBagFoldHigh\times$ vs.\ ERM, and
twin-bootstrap reduces it by another factor of
${\sim}\symKLTwinAdditionalFoldMedian$ (per-dataset paired CIs in
Table~\ref{tab:distributional}, Appendix~\ref{app:distributional}).
Pipelines that consume the softmax --- active-learning acquisition,
Bayesian-optimisation, virtual-screening top-$K$ ranking --- see a
larger gap between methods than argmax churn reflects.
Appendix~\ref{app:bo_topk} quantifies this directly: twin-bootstrap
raises the Jaccard overlap of the top-$\boTopkK$ predicted-active
sets between two retrainings on every chemistry dataset (paired
$\Delta J_{\boTopkK} \geq \boTopkTwinDeltaMin$ on every cell), with
the largest gains where ERM is most unstable.  In a full BO-loop
simulation on the three regression benchmarks, twin-bootstrap
reduces the cross-trajectory standard deviation of the final-best
acquired $y$ by
$\boLoopRegTwinStdRedLow\%$--$\boLoopRegTwinStdRedHigh\%$ on every
dataset; bagging-$K{=}5$ helps on
$\boLoopRegBagStdRedWins/\boLoopRegNumDatasets$ but is comparable
to ERM on the third (Appendix~\ref{app:bo_loop_reg}).

\begin{figure}[t]
  \centering
  \includegraphics[width=0.75\linewidth]{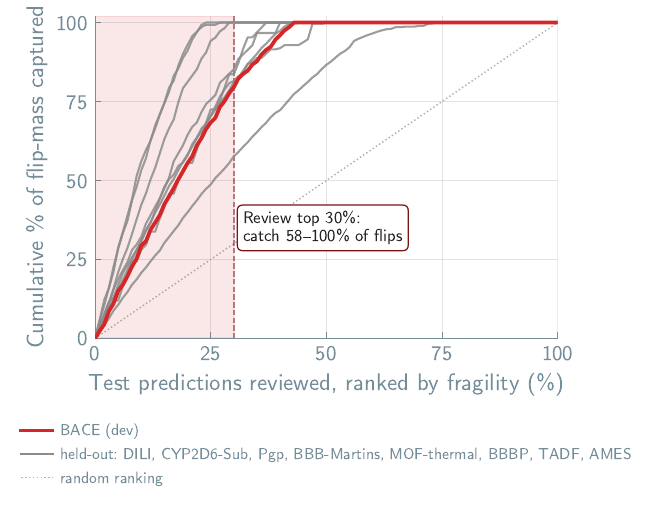}
  \caption{\textbf{\boldmath Routing the top $\reviewFraction\%$ of test predictions
    (ranked by per-example churn from one extra retraining) to a
    human reviewer captures $\triageKtenLow$--$\triageKtenHigh\%$ of
    all retraining-induced class flips on the \nDatasetsHeadline{}
    chemistry datasets.}  Each curve is one
    dataset (BACE in red, held-out in grey); $x$-axis is the fraction
    of test predictions reviewed in churn-rank order, $y$-axis
    is the cumulative fraction of total flip-mass captured.  The
    diagonal marks random ranking; the shaded band marks the
    practical review fraction.  The triage workflow is a
    one-bootstrap operation on top of an existing pipeline.}
  \label{fig:decile}
\end{figure}

\paragraph{Churn-ranked triage.}
\label{sec:churn-decile}
Per-example churn, computed across pairs of ERM bootstraps, ranks
test examples by how likely they are to flip class on retraining.
Figure~\ref{fig:decile} reports the cumulative fraction of total
flip-mass captured if the practitioner sorts predictions by
churn and reviews the top fraction.  Reviewing the top $\reviewFraction\%$ captures
$\triageKtenLow\text{--}\triageKtenHigh\%$ of all flips when churn
is scored from all $10$ bootstraps.  With a single extra retraining
($K{=}2$, the one-bootstrap practitioner workflow), the
top-$\reviewFraction\%$ recall drops only to
$\triageKtwoLow\text{--}\triageKtwoHigh\%$ -- a gap of just
$\triageKtwoToKtenGapLow\text{--}\triageKtwoToKtenGapHigh$\,pp
from the $K{=}10$ gold standard on every dataset
(Appendix~\ref{app:convergence}).  Predictive entropy from a single
ERM model ranks worse on this task: churn captures
$\fragTopLow\text{--}\fragTopHigh\%$ of flips in its top decile vs.\
$\entropyTopLow\text{--}\entropyTopHigh\%$ for entropy (per-dataset
gap $\entropyGapMin\text{--}\entropyGapMax$\,pp,
Appendix~\ref{app:entropy}).
A practitioner can therefore identify the $\topDecilePct\%$ most
fragile predictions from a single additional retraining: one extra
bootstrap on top of the deployed pipeline.

A per-molecule view of these flips on six BACE id-test compounds is
in Appendix~\ref{app:case_study} (Figure~\ref{fig:case_study}).

\subsection{The recipe generalises across architectures and tasks}
\label{sec:generalisation}

The main result on small-$N$ from-scratch MLPs holds across three
other architectures and on regression.  Different architectures
require different $\lambda$ values, but the same accuracy-tolerance
rule on the development set (Section~\ref{sec:methods},
``Hyperparameter selection'') picks each one.

\paragraph{Pretrained backbones.}
The $\lambda{=}300$ chosen on the BACE MLP over-constrains pretrained
representations: twin-bootstrap collapses accuracy by
$\waterbirdsAccCollapse$pp on Waterbirds (ImageNet ResNet-50) and
by $\chembertaAccDropLow\text{--}\chembertaAccDropHigh$pp on
ChemBERTa-77M-MTR.  Re-applying the
$\preregTolerance$-tolerance selection rule on each backbone's
development set picks $\lambda{=}10$, which preserves accuracy and
cuts the class-flip rate on every ChemBERTa dataset
(Appendix~\ref{app:chemberta_scope}) and on Waterbirds
(Appendix~\ref{app:waterbirds_lambda}).  Bagging is independent of
$\lambda$ and works on every backbone we tested.

\paragraph{Graph networks.}
On a $3$-layer GIN on RDKit graphs of BACE, bagging-$K{=}5$ transfers
cleanly and improves id-accuracy by $\ginBagAccGain$pp.  Twin-bootstrap
at $\lambda{=}300$ collapses accuracy and is rejected by the same
$\preregTolerance$-tolerance rule; re-running the rule on BACE-GIN
selects $\lambda{=}\ginRuleLam$, which cuts the class-flip rate
$\ginCutLamTen\%$ at no accuracy cost (Appendix~\ref{app:gin}).

\paragraph{Regression.}
On three regression benchmarks (ESOL, FreeSolv, Lipophilicity;
Appendix~\ref{app:regression}) the methods extend to continuous
targets and rank the same way as on classification.  Cross-sample
churn for regression is the per-example absolute prediction
difference between two retrainings; on ERM, that churn is
$\regFragRatioLow\text{--}\regFragRatioHigh\%$ of the deployed
model's mean absolute error --- a single retraining moves predictions
by roughly half the test-set MAE on average.  At matched
$2\times$-ERM compute, twin-bootstrap cuts prediction-difference
churn $\regTwinLow\text{--}\regTwinHigh\%$ vs.\ ERM while
bagging-$K{=}2$ cuts $\regBagTwoLow\text{--}\regBagTwoHigh\%$
(twin-bootstrap wins on every dataset).  Bagging-$K{=}5$ at
$5\times$-ERM compute is strongest at
$\regBagFiveLow\text{--}\regBagFiveHigh\%$.

%% file: sections/tables/main.tex
\begin{table}[t]
  \centering
  \caption{\textbf{Bagging and twin-bootstrap reduce the class-flip rate on every chemistry benchmark; MC dropout and deep ensembles do not.}  Paired $\Delta$ id-churn vs.\ ERM in percentage points (negative is better).  Each cell is the mean across $\nCanonicalSeeds$ canonical-seed replicates; each replicate-mean averages over the $45$ pairs that $10$ retrainings produce, with paired-bootstrap $95\%$ CIs from $10{,}000$ resamples.  Per-replicate values in Appendix~\ref{app:seed_sensitivity}.  \textbf{Best} per dataset in bold; entries whose CI excludes zero are significant at $\alpha{=}0.05$.  Twin-bootstrap $\lambda{=}300$ is selected on BACE only and applied unchanged to every held-out benchmark.}
  \label{tab:main}
  \scriptsize
  \begin{tabular}{lrll@{\hspace{1.2em}}lll}
    \toprule
    & & \multicolumn{2}{c}{Parameter-side} & \multicolumn{3}{c}{Data-side} \\
    \cmidrule(lr){3-4} \cmidrule(lr){5-7}
    Dataset & $N$ & MC dropout & Deep Ens.\ $K{=}5$ & Bagging $K{=}2$ & Bagging $K{=}5$ & Twin-bootstrap $\lambda{=}300$ \\
    \midrule
    DILI & 304 & +1.1 [+0.3, +1.8] & +0.2 [-0.6, +1.1] & -3.9 [-5.4, -2.5] & \textbf{-8.4 [-9.8, -7.0]} & -2.7 [-4.2, -1.1] \\
    CYP2D6-Sub & 427 & +0.0 [-0.4, +0.4] & +0.1 [-0.5, +0.7] & -6.0 [-7.0, -5.1] & -7.2 [-8.2, -6.2] & \textbf{-9.3 [-10.2, -8.5]} \\
    Pgp & 780 & -0.2 [-0.6, +0.1] & +0.3 [-0.1, +0.6] & -2.3 [-3.2, -1.5] & \textbf{-4.4 [-5.1, -3.7]} & -3.6 [-4.3, -3.0] \\
    BACE\,(dev) & 968 & +0.3 [-0.2, +0.7] & -0.3 [-0.7, +0.1] & -2.8 [-3.6, -2.0] & -6.5 [-7.2, -5.7] & \textbf{-10.5 [-11.0, -9.9]} \\
    TADF & 1007 & -0.5 [-0.8, -0.1] & -0.5 [-1.0, +0.0] & -3.0 [-3.5, -2.5] & \textbf{-6.0 [-6.5, -5.4]} & -4.7 [-5.4, -4.0] \\
    MOF-thermal & 1251 & -1.6 [-2.4, -0.9] & -4.6 [-5.3, -3.9] & -5.8 [-6.6, -5.0] & -10.5 [-11.2, -9.7] & \textbf{-14.1 [-14.9, -13.3]} \\
    BBB-Martins & 1300 & +1.0 [+0.7, +1.3] & -0.4 [-0.7, -0.2] & -2.7 [-3.2, -2.3] & -4.1 [-4.5, -3.7] & \textbf{-7.4 [-7.8, -7.0]} \\
    BBBP & 1305 & +0.6 [+0.3, +0.9] & -0.6 [-0.9, -0.4] & -2.0 [-2.5, -1.4] & -4.0 [-4.5, -3.6] & \textbf{-7.0 [-7.5, -6.6]} \\
    AMES & 4658 & +1.8 [+1.5, +2.1] & -0.5 [-0.7, -0.3] & -3.9 [-4.3, -3.6] & -7.2 [-7.6, -6.8] & \textbf{-8.3 [-8.6, -7.9]} \\
    \bottomrule
  \end{tabular}
\end{table}

%% file: sections/scope.tex
\section{Limitations}
\label{sec:limitations}

The main magnitudes and method comparison are on chemistry binary
classification (MoleculeNet, TDC ADME and Tox, materials-science).
Regression is covered for three MoleculeNet benchmarks
(Appendix~\ref{app:regression}); Waterbirds is a single vision
cross-check; multi-class classification and structured outputs are
not tested.  We use $K{=}2$ throughout twin-bootstrap and do not study larger head
counts.  

We measure cross-sample churn at the prediction layer, at the
top-$\boTopkK$ ranking layer that gates a single
Bayesian-optimisation acquisition step
(Appendix~\ref{app:bo_topk}), and at the trajectory level on the
three regression benchmarks (Appendix~\ref{app:bo_loop_reg}:
twin-bootstrap reduces the cross-trajectory std of the final-best
$y$ by $\boLoopRegTwinStdRedLow\%$--$\boLoopRegTwinStdRedHigh\%$ in
a $\boLoopRegBudget$-step greedy BO loop on every dataset; bagging
on $2/3$).  We use greedy top-$1$ acquisition only; UCB, expected
improvement, and Thompson sampling are not tested.

%% file: sections/discussion.tex
\section{Discussion}
\label{sec:discussion}

\paragraph{Cross-sample churn is a metric scientific-ML benchmarks have been missing.}
Parameter-side uncertainty methods that the field treats as adequate
--- deep ensembles, MC dropout, stochastic weight averaging ---
do not reduce cross-sample churn on these benchmarks, and on several
they make it worse.  This follows from how they are constructed:
they sample over weights at fixed data, while the variance
practitioners act on lives on the data axis they hold constant.  A
performance-only benchmark cannot separate them from data-side
methods that do reduce the rate, because the per-prediction signal
they differ on is not part of the standard report.  Adding
cross-sample churn to the standard report exposes the methods that
produce different per-prediction calls between retrainings even when
aggregate accuracy looks the same --- the regime where the model's
choice of next molecule is bootstrap-sensitive and the workflow
built on top of it is not reproducible.

\paragraph{The data-resampling protocol is the design lever.}
Cross-sample stability is set less by the model class than by how
the training procedure resamples data between independent runs.  The
codistillation-vs.\ twin-bootstrap comparison
(Appendix~\ref{app:overlap_spectrum}) isolates the inter-network
bootstrap overlap as the operative parameter; the consistency weight
$\lambda$ that controls how strongly the procedure exploits the
overlap takes a different numerical value on different
architectures, but the development-set rule that selects it
transfers unchanged across MLP, GIN, ChemBERTa, and ImageNet-pretrained
ResNet-50 (Section~\ref{sec:generalisation}).  Cross-sample stability
can therefore be designed in at training time, without changing the
model class or the deployment pipeline, by tuning a single
hyperparameter on the development set with the same rule.

\paragraph{Broader impacts.}
Reducing cross-sample churn has operational consequences because
deployment-driven domains route substantial wet-lab effort to the
molecules a model ranks highest at any given retraining; reducing
the rate at which that ranking flips between dataset updates cuts
wasted experimental work and makes computational triage decisions
reproducible.  However, low churn is not correctness: a stably wrong
model looks identical to a stably right
one under this metric, and the churn-ranked triage workflow only
flags predictions that disagree across retrainings, not those that
agree but are jointly miscalibrated.  Cross-sample churn is also a
bootstrap-variance quantity --- it does not characterise stability
under distribution shift between the training population and the
deployment domain, which requires separate measurement.

%% file: sections/appendix.tex
\noindent\textbf{Appendix roadmap.}
Appendices~\ref{app:nscaling}--\ref{app:filter_outcomes} support the
measurement framework ($N$-scaling, per-class breakdown,
aggregate-metric drift, borderline datasets, and the
ERM-vs-majority filter outcomes).
Appendix~\ref{app:seed_sensitivity} reports the canonical-seed
sensitivity of every magnitude in the main table.
Appendices~\ref{app:pareto}--\ref{app:ablations} support the methods
(Pareto curve, twin-bootstrap bayesian optimisation, overlap
spectrum, ablations).
Appendices~\ref{app:case_study}--\ref{app:bo_loop_reg} support the
empirical results (per-molecule case study, distributional
disagreement, Friedman test, triage convergence, entropy baseline,
top-$K$ ranking stability and a full BO-loop simulation on regression
as Bayesian-optimisation analogues).
Appendices~\ref{app:gin}--\ref{app:regression} document the
cross-architecture and cross-task generalisation evidence.
Appendix~\ref{app:compute} reports the per-method compute footprint.

\section{\boldmath$N$-scaling on BACE}
\label{app:nscaling}

The cross-dataset variation in Table~\ref{tab:fragility-magnitudes}
confounds training-set size with task difficulty.  We isolate $N$ by
subsampling a single dataset (BACE) to
$M \in \{200, 300, 400, 500, 600, 700, 800, 900, 968\}$ and re-running
ERM at each $M$ for $\nSeeds$ train-seeds; the pool at each $M$ is the
deterministic prefix of a fixed canonical-seed shuffle, so all
$\nSeeds$ train-seeds at the same $M$ draw their bootstraps from the
same underlying pool.  Sym-KL trends downward as $M$ grows: the
seed-averaged value falls from $\nScalingSymKLLow$ at $M{=}200$ to
$\nScalingSymKLMin$ at $M{=}\nScalingMmin$ before ticking back up
slightly to $\nScalingSymKLEnd$ at $M{=}968$ (Table~\ref{tab:nscaling};
log-log slope $\nScalingSlope$, computed on canonical-seed
$\canonicalSeed$ alone for protocol consistency).  More data reduces
churn but does not drive it to zero, motivating the methods we
propose for the regime where simply collecting more data is not
available.

\input{sections/tables/nscaling_bace.tex}

\paragraph{Sensitivity to the canonical-pool draw.}  The
canonical-seed shuffle determines which examples enter the pool at
each $M$, so the per-$M$ sym-KL is itself a single-seed estimate.  We
re-ran the entire $M$-grid on two additional canonical seeds ($7$ and
$42$) for $\nSeeds$ train-seeds each; the per-seed sym-KL at fixed $M$
ranges by up to $\sim$$30\%$ in absolute terms, but the averaged
trajectory preserves the broad-decreasing shape described above.
The slight uptick from $M{=}\nScalingMmin$ to $M{=}968$ is
canonical-seed sensitive (visible on canonical seeds $\canonicalSeed$
and $7$, absent on $42$) and should be read as a soft floor near
$M{\approx}\nScalingMmin$ rather than a sharp inflection.

\section{Per-class breakdown: minority-class predictions are more unstable}
\label{app:per_class_churn}

The overall cross-bootstrap argmax-churn rate is an average over the
canonical id-test set.  On imbalanced datasets, that average can hide
class-conditional structure: are the disagreements concentrated on
majority-class examples (where the model is essentially predicting
the dominant prior) or on minority-class examples (where the model
has to learn structure)?  Table~\ref{tab:per-class-churn} restricts
churn to each true-label subset.  On the three most imbalanced
chemistry datasets in our suite (BBB-Martins, BBBP at positive
fraction $\imbalPosFracHigh$; CYP2D6-Sub at $\cypTwoPosFrac$), minority-class predictions
are $2\text{--}4\times$ more unstable than majority-class
predictions.  The disagreement is therefore concentrated on the
class practitioners care most about: the predicted-positive (active,
permeable, substrate) calls that drive next-step decisions.

\input{sections/tables/per_class_churn.tex}

\section{\boldmath Aggregate-metric drift under precision, recall, $F_1$, AP}
\label{app:additional_metrics}

Table~\ref{tab:fragility-magnitudes} reports
$|\Delta\text{accuracy}|$ between two retrainings.  Accuracy on
imbalanced binary tasks can mask class-specific drift, so we
additionally compute paired $|\Delta|$ for precision, recall, $F_1$,
and average precision (AP) on each of the \nDatasetsHeadline{} chemistry
datasets, using the same $\nSeedPairs$ seed pairs of ERM
bootstraps.
Table~\ref{tab:additional-metrics} reports the result.  On the
imbalanced ADME datasets (CYP2D6-Sub, DILI), recall and
precision drifts are $2\text{--}3\times$ accuracy drift, but the
per-example argmax-churn rate ($\cypTwoOverallChurn\%$ on CYP2D6-Sub, $\diliOverallChurn\%$ on DILI)
still exceeds every aggregate-metric drift on every dataset.

\input{sections/tables/additional_metrics.tex}

\section{Borderline datasets}
\label{app:borderline}

\paragraph{hERG, HIA\_Hou, Skin\_Reaction.}
The headline analysis requires ERM to clear majority-class
accuracy by at least $\preregFilterPP$\,pp, otherwise cross-sample
churn would conflate the model shifting its decision boundary with
the majority class itself shuffling under sampling noise.  These
three datasets pass that filter only marginally
($+\borderlineGapMin$ to $+\borderlineGapMax$\,pp on test sets of
$\borderlineNTestMin$--$\borderlineNTestMax$ examples).  Their
cross-sample magnitudes are reported below for transparency; the
method comparison is not run on them because the small test sets do
not give enough statistical power.

\input{sections/tables/borderline_magnitudes.tex}

\section{Filter outcomes for excluded datasets}
\label{app:filter_outcomes}

Table~\ref{tab:filter-outcomes} reports the per-dataset ERM id-acc
and the canonical majority-class baseline on the five chemistry
datasets that fail the $+\preregFilterPP$\,pp filter and are
therefore excluded from the main analysis.

\input{sections/tables/filter_outcomes.tex}

\section{Canonical-seed sensitivity}
\label{app:seed_sensitivity}

The canonical seed determines the train/test split.  All
single-canonical-seed estimates of cross-sample churn therefore
inherit a sensitivity to which examples ended up in the test set.
We replicate the entire main-table protocol on $\nCanonicalSeeds$
canonical seeds ($\canonicalSeed$, $7$, $42$) for each method-dataset
cell ($\nSeeds$ train-seeds each), and report the per-cell across-seed
mean (Table~\ref{tab:main}, Table~\ref{tab:fragility-magnitudes}) with
the per-seed values tabulated below.

\paragraph{What changes across canonical seeds.}
The largest across-seed range observed for any single
(dataset, method) cell is $\maxAcrossSeedChurnSpreadPP$\,pp on the
absolute class-flip rate, attained on \maxAcrossSeedChurnSpreadCell.
Magnitudes are seed-sensitive because the test set composition
changes between canonical seeds.  Paired comparisons --- the main
$\Delta$-churn columns of Table~\ref{tab:main} --- are computed within
seed and so factor out the test-set composition; the largest
across-seed range of any paired $\Delta$ in our suite is
$\maxAcrossSeedDeltaSpreadPP$\,pp (\maxAcrossSeedDeltaSpreadCell),
typically a small fraction of the magnitude of the reduction.

\paragraph{What does not change.}
The qualitative ranking ``data-side $>$ parameter-side'' (bagging and
twin-bootstrap reduce churn vs.\ ERM on every seed; deep ensemble /
MC dropout / SWA do not consistently reduce churn) holds on every
canonical seed individually.  Wins counts and Friedman ranks computed
per-seed agree with the across-seed average reported in
Section~\ref{sec:experiments}.  The $\preregTolerance$-tolerance
$\lambda$-selection rule on BACE selects $\lambda{=}300$ on every
canonical seed.

\input{sections/tables/seed_sensitivity.tex}

\paragraph{Pretrained-backbone scope studies.}
For the pretrained-backbone scope studies (GIN, ChemBERTa, Waterbirds:
Appendices~\ref{app:gin}--\ref{app:waterbirds_lambda}) we replicate
ERM and the rule-selected $\lambda$ on all $\nCanonicalSeeds$
canonical seeds; the development $\lambda$-Pareto sweeps within those
studies are run on canonical seed $\canonicalSeed$ only, since the
goal there is architecture-feasibility evidence and the rule-selected
$\lambda$ does not change across seeds.

\section{Development Pareto curve on BACE}
\label{app:pareto}

\begin{figure}[t]
  \centering
  \includegraphics[width=0.8\linewidth]{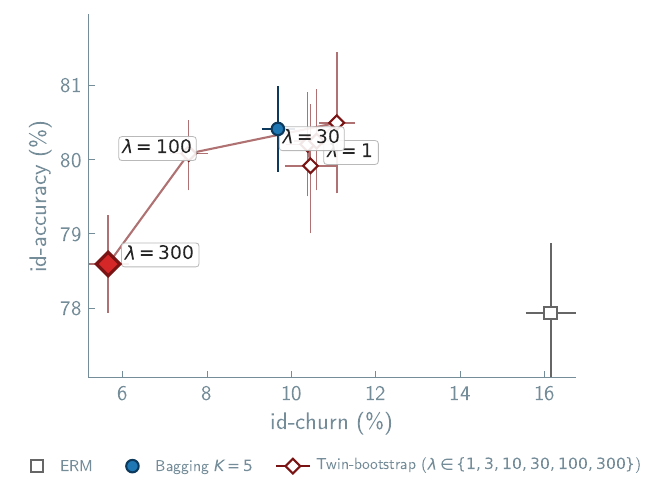}
  \caption{\textbf{Twin-bootstrap $\lambda{=}300$ sits at the
    accuracy-preserving end of the BACE Pareto frontier.}  Twin-bootstrap
    at $\lambda \in \{1, 3, 10, 30, 100, 300\}$ traces an
    accuracy-vs.-churn trajectory; the pre-registered selection rule
    (largest $\lambda$ with id-acc $\ge$ ERM-id-acc $-\preregTolerance$) picks
    $\lambda{=}300$ (filled red).  Bagging-$K{=}5$ (blue circle)
    achieves a similar accuracy at higher churn; ERM (open square,
    upper-right) at the highest churn.  Error bars are bootstrap
    $95\%$ CIs.}
  \label{fig:pareto}
\end{figure}

A plateau of configurations $\lambda \in \{1, 3, 10, 30\}$ all preserve
accuracy and reduce churn from ERM's $\baceErmChurn\%$ to a low-single-digit
range; the curve drops further at higher $\lambda$ until
$\lambda{=}300$ at $\baceTwinChurn\%$ within the $\preregTolerance$ accuracy
tolerance (Figure~\ref{fig:pareto}).  The selected $\lambda$ sits at the extreme of the feasible interval
that satisfies the $\preregTolerance$ accuracy tolerance, and the
choice is robust to the exact tolerance value: relaxing the tolerance
to $0.01$ would still pick the same $\lambda$, while tightening to
$0.005$ would pick $\lambda{=}100$ (still on the knee).

\paragraph{Accuracy at the frozen $\lambda$ across held-out datasets.}
At the frozen $\lambda{=}300$, twin-bootstrap improves id-accuracy on
$3/9$ chemistry datasets (BACE $+1.2$\,pp, DILI $+2.2$\,pp, TADF
$+0.5$\,pp) and is within paired-bootstrap CIs on AMES, BBB-Martins,
CYP2D6-Sub, and BBBP.  Pgp shows a small loss ($-1.5$\,pp), and
MOF-thermal collapses ($-7.1$\,pp, well outside the $0.02$ selection-rule
tolerance and the only dataset on which a per-dataset re-application of
the rule would reject the BACE-frozen $\lambda$).

\section{\boldmath Per-dataset Bayesian optimisation of $\lambda$}
\label{app:bayes_twin}

The pre-registered rule (Section~\ref{sec:methods}) freezes
$\lambda{=}\preregLambda$ on BACE and applies it to every other
dataset.  Does per-dataset tuning win on the metric the paper reports
(cross-sample churn between independent retrainings)?  We test this
with a Bayesian optimisation that minimises the argmax disagreement
between two independent twin-bootstrap ensembles on a held-out
validation fold, subject to an accuracy constraint relative to the
unregularised baseline -- the val-side analogue of the cross-sample
churn the paper reports on test.

\paragraph{Protocol.}
For each (dataset, training seed), $\bayesTwinTrials$ BO trials over
$\log_{10}\lambda \in [\log_{10}\bayesTwinLamMin,\,
\log_{10}\bayesTwinLamMax]$.  Each trial trains \emph{two}
independent twin-bootstrap pairs on the same fold-train pool with
different training seeds, and measures the argmax disagreement
between their two ensembles on the held-out fold
($\text{val\_cross\_churn}$).  The first trial is forced to
$\lambda{=}0$ so the unregularised baseline accuracy $a_0$ is set
before the constraint kicks in; the next $\bayesTwinInitTrials$ trials
are log-uniform random, after which BO acquisition uses GP-EI
($\bayesTwinKernel$ kernel) on the scalar
\[
  \text{score}(\lambda) = -\text{val\_cross\_churn}(\lambda)
    - C\cdot\max\bigl(0,\, a_0 - \preregTolerance - \text{val\_acc}(\lambda)\bigr).
\]
With $\text{val\_cross\_churn} \in [0,1]$ and $C{=}\churnPenalty$,
an accuracy shortfall of $0.01$ subtracts $1.0$ from the score, i.e.\
the full feasible-region range of the first term -- so BO is pulled
back inside the accuracy constraint within a few trials of any
violation.  Val scores average over $k{=}3$ cross-validation folds
carved deterministically from the canonical training pool with seed
$\canonicalSeed$; \texttt{id\_test} and \texttt{ood\_test} are never
seen during selection.  Algorithm~\ref{alg:bayes_twin_bo} gives the
loop; we denote a twin-bootstrap pair as $\theta_A, \theta_B$ (two
networks trained jointly on independent bootstraps with the sym-KL
consistency loss; deployment averages their softmax outputs).

\begin{algorithm}[H]
  \caption{Per-dataset BO of $\lambda$ with cross-sample-churn
           objective.  Stage~1 (the outer \textbf{for}-loop over $s$)
           runs once per (dataset, training seed) and returns
           $\lambda^\star_s$; stage~2 (the two trailing lines)
           takes the median across seeds and retrains.}
  \label{alg:bayes_twin_bo}
  \begin{algorithmic}[1]
    \Require fold-train indices $\{I_f\}_{f=1}^{k}$ and val loaders
             $\{V_f\}_{f=1}^{k}$; trial count $T$;
             init-trial count $T_0$;
             accuracy tolerance $\delta$; penalty $C$;
             training seeds $\{s_1, \dots, s_S\}$
    \For{$s \in \{s_1, \dots, s_S\}$}
      \Comment{stage 1: per-seed BO}
      \State $\lambda_1 \gets 0$ \Comment{forced baseline trial}
      \For{$t = 1, \dots, T$}
        \State $\bar a, \bar c \gets 0, 0$
        \For{$f = 1, \dots, k$}
          \State $(\theta_A, \theta_B) \gets$ twin-bootstrap on $I_f$ at $\lambda_t$
          \State $(\theta_A^\prime, \theta_B^\prime) \gets$ twin-bootstrap on $I_f$ at $\lambda_t$
                 with an independent training seed
          \State $p \gets \tfrac{1}{2}(f_{\theta_A} + f_{\theta_B})$,
                 \, $p^\prime \gets \tfrac{1}{2}(f_{\theta_A^\prime} + f_{\theta_B^\prime})$
                 on $V_f$
          \State $\bar a \mathrel{+}= \text{acc}(p, V_f) / k$
          \State $\bar c \mathrel{+}= \Pr_{x \in V_f}[\arg\max p(x) \neq \arg\max p^\prime(x)] / k$
        \EndFor
        \If{$t = 1$} $a_0 \gets \bar a$ \EndIf
        \State $\text{score}_t \gets -\bar c - C \cdot \max(0, a_0 - \delta - \bar a)$
        \State $\lambda_{t+1} \gets$ log-uniform random if $t \leq T_0$, else GP-EI argmax over
               $\{(\log_{10}\lambda_i, \text{score}_i)\}_{i \leq t}$
      \EndFor
      \State $\lambda^\star_s \gets \lambda_{\arg\max_t \text{score}_t}$
    \EndFor
    \State $\lambda_{\text{dataset}} \gets \mathrm{median}\bigl(\lambda^\star_{s_1}, \dots, \lambda^\star_{s_S}\bigr)$
           \Comment{stage 2: aggregate}
    \For{$s \in \{s_1, \dots, s_S\}$}
      \State retrain twin-bootstrap on the full canonical pool at $\lambda_{\text{dataset}}$
             with training seed $s$
    \EndFor
    \State \Return the $S$ retrained twins
  \end{algorithmic}
\end{algorithm}

Cross-sample churn between any two retrainings within a dataset is
then fixed-$\lambda$ variance on both arms of the comparison vs.\
the frozen $\lambda{=}\preregLambda$ baseline.  The full protocol
(per-seed BO sweep, median aggregation, fixed-$\lambda$ retraining,
and table regeneration) is reproducible with the repository's
\texttt{make bayes-twin} target.

\paragraph{Findings.}
The per-dataset tuned $\lambda$ varies by roughly two orders of
magnitude and lands on both sides of $\preregLambda$
(Table~\ref{tab:bayes_twin}, ``Mean $\lambda$'' column).  On
cross-sample ID-churn -- the metric Table~\ref{tab:main} reports --
per-dataset BO beats the rule on
$\bayesTwinBeatsRuleCount/\nDatasetsHeadline$ datasets (CI excludes
zero), ties on $\bayesTwinTiesRuleCount/\nDatasetsHeadline$, and
loses on $\bayesTwinLosesRuleCount/\nDatasetsHeadline$.  Against
bootstrap-matched ERM (Table~\ref{tab:main}), per-dataset BO cuts
cross-sample ID-churn on every dataset, so per-dataset tuning
preserves and modestly extends the twin-bootstrap effect rather than
erasing it.

The frozen $\lambda{=}\preregLambda$ is therefore a robust default --
it matches or beats per-dataset BO on
$\bayesTwinTiesOrLosesCount/\nDatasetsHeadline$ datasets -- but not
the per-dataset optimum everywhere: per-dataset tuning is worth the
BO cost when the dataset's class structure or size diverges from
BACE.

\input{sections/tables/bayes_twin.tex}

\section{\boldmath Overlap spectrum: $\sim$\overlapPctMid\% balances churn reduction and accuracy preservation}
\label{app:overlap_spectrum}

The codistillation~\citep{anil2018large},
twin-bootstrap, and twin-shared variants share the same two-network
sym-KL consistency objective and differ only in the inter-network
bootstrap-overlap: $\overlapPctLo\%$ (disjoint shards),
${\sim}\overlapPctMid\%$ (independent bootstraps with replacement),
and $\overlapPctHi\%$ (the same bootstrap to
both networks).  Table~\ref{tab:overlap_spectrum} reports paired
$\Delta$ class-flip rate vs.\ ERM at each operating point on every dataset
where all three methods completed.

\input{sections/tables/overlap_spectrum.tex}

Codistillation has the largest mean churn reduction on most datasets
but drops id-accuracy by more than $\accDegradeFlagPP$\,pp on Pgp and
MOF-thermal; twin-shared drops accuracy on MOF-thermal and increases
churn on DILI; twin-bootstrap at ${\sim}40\%$ matches codistillation
on never increasing churn while only collapsing accuracy on a single
dataset (MOF-thermal, where all three operating points collapse).  Among the three points, only twin-bootstrap at
${\sim}\overlapPctMid\%$ overlap satisfies both criteria
simultaneously, which is what motivates its use as the default
operating point.

\section{Weight-decay ablation}
\label{app:ablations}

\paragraph{Weight decay alone is insufficient.}
Sweeping AdamW weight decay over five orders of magnitude
($10^{-5}$ to $10^{-1}$) on BACE, BBBP, and TADF produced no detectable
movement in cross-sample churn (paired CIs vs.\ ERM include zero at
every value tested).  The effective per-step regularisation under
AdamW does not constrain the data-resampling sensitivity that drives
churn.

\section{Per-molecule case study on BACE}
\label{app:case_study}

\begin{figure}[h]
  \centering
  \includegraphics[width=\linewidth]{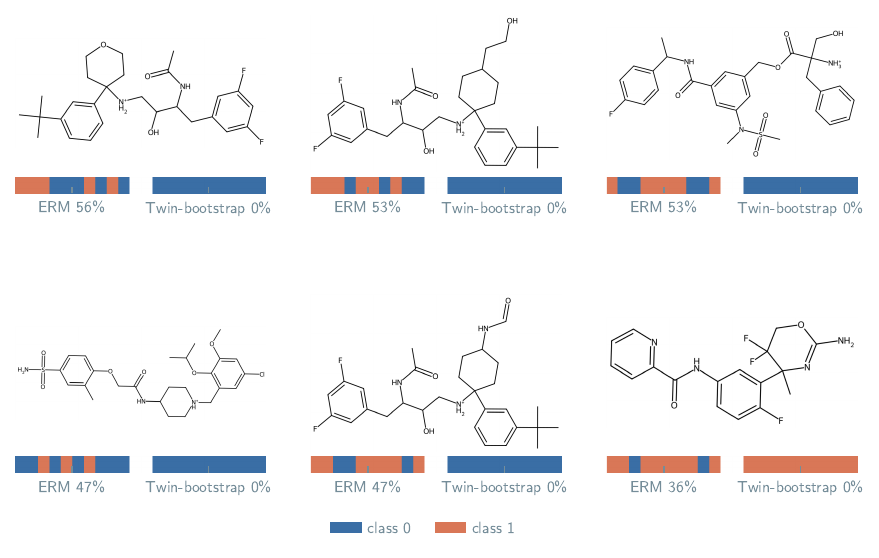}
  \caption{\textbf{Twin-bootstrap eliminates per-molecule retraining
    flips on BACE.}  Six id-test molecules where ERM flips class on
    $\geq 36\%$ of seed pairs and twin-bootstrap flips on $0\%$.
    Below each structure: ten ERM and ten twin-bootstrap seed
    predictions, each cell is the predicted class at one retraining;
    the gap separates the two methods.  ERM stripes flip;
    twin-bootstrap collapses to a single colour.}
  \label{fig:case_study}
\end{figure}

Figure~\ref{fig:case_study} shows six BACE id-test molecules where
ERM flips class on $\geq 36\%$ of seed pairs and twin-bootstrap flips
on $0\%$ over the same ten retrainings.  Both methods see the same
canonical training pool and test set; the difference is the
consistency loss.

\section{Distributional disagreement (sym-KL) per dataset}
\label{app:distributional}

Table~\ref{tab:distributional} reports paired $\Delta$ sym-KL vs.\
ERM per dataset and per method, in the same format as the
argmax-churn main result (Table~\ref{tab:main}).  Aggregating over
datasets, bagging-$K{=}5$ cuts sym-KL by
$\symKLBagFoldLow\text{--}\symKLBagFoldHigh\times$ vs.\ ERM and
twin-bootstrap by a further median ${\sim}\symKLTwinAdditionalFoldMedian\times$;
the table below shows the per-method, per-dataset paired CIs behind
those aggregates.

\input{sections/tables/distributional.tex}

\section{\boldmath Friedman test: twin-bootstrap matches bagging-$K{=}5$ in mean rank}
\label{app:friedman}

The pairwise CIs reported in Table~\ref{tab:main} test each
(method, dataset) cell separately.  As an aggregate cross-dataset
test, we apply the Friedman non-parametric rank test on mean
cross-sample class-flip rate over the nine datasets and seven methods (ERM,
MC dropout, SWA, Deep Ensemble $K{=}5$, Bagging $K{=}2$, Bagging
$K{=}5$, Twin-bootstrap $\lambda{=}300$).  The test rejects the null
of equal ranks at $\chi^2 = \friedmanChi$, $p = \friedmanP$.  Mean
ranks (lower is better, $1$ = best on cross-sample class-flip rate) are
twin-bootstrap $\friedmanRankTwin$, bagging $K{=}5$
$\friedmanRankBagFive$, bagging $K{=}2$ $\friedmanRankBagTwo$, SWA
$\friedmanRankSWA$, deep-ensemble $K{=}5$ $\friedmanRankDeepEns$,
ERM $\friedmanRankERM$, MC dropout $\friedmanRankMCD$.  The Nemenyi
critical difference at $\alpha{=}0.05$ is
$\textit{CD}{=}\friedmanCD$; rank gaps exceeding $\textit{CD}$
identify pairs that differ significantly.  Twin-bootstrap is
significantly better in rank than ERM, MC dropout, SWA, and deep
ensemble; it is \emph{not} significantly different from bagging
$K{=}5$ (rank gap $\friedmanRankGapBagFiveTwin$).  Twin-bootstrap therefore matches bagging
$K{=}5$ in mean-rank at roughly $2{\times}$ ERM compute
(Appendix~\ref{app:compute}) versus the $5{\times}$ that bagging
$K{=}5$ requires.

\section{Convergence: how many bootstraps suffice for triage?}
\label{app:convergence}

The triage workflow asks the practitioner to train one extra ERM
bootstrap and rank predictions by churn.  We verify that
$K{=}2$ (one extra bootstrap) is sufficient by comparing the
top-$\reviewFraction\%$ cumulative recall when the churn ranking is computed
from $K \in \{2, 3, 5, 10\}$ ERM bootstraps.

\begin{figure}[h]
  \centering
  \includegraphics[width=0.75\linewidth]{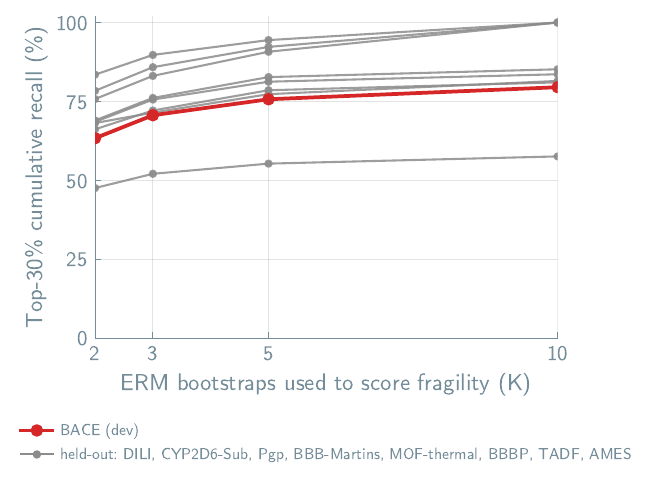}
  \caption{\textbf{One extra bootstrap is enough: the top-$\reviewFraction\%$
    recall at $K{=}2$ is within $\triageKtwoToKtenGapLow$--$\triageKtwoToKtenGapHigh$\,pp of the $K{=}10$ gold
    standard on every dataset.}  Mean recall across $30$ random
    $K$-subsets per $K{<}10$.  Per-dataset at $K{=}2$:
    $\triageKtwoLow$--$\triageKtwoHigh\%$.  At $K{=}10$ (using all
    ten bootstraps to score churn):
    $\triageKtenLow$--$\triageKtenHigh\%$.  MOF-thermal is the floor
    on every $K$ ($\mofThermalConvLow$--$\mofThermalConvHigh\%$);
    BBBP and BBB-Martins reach $\overlapPctHi\%$ recall by $K{=}5$.}
  \label{fig:convergence}
\end{figure}

\section{Predictive entropy: a weaker flip-predictor than churn on every dataset}
\label{app:entropy}

A natural deployment-cheaper alternative to per-example churn is
the predictive entropy of a single trained model: the practitioner
already has access to a softmax distribution at deployment time and
need not train a second model.  We test whether the per-example
entropy ranks test predictions for retraining-flip risk as well as
churn does.

\input{sections/tables/entropy_vs_fragility.tex}

Across the eight held-out datasets, churn captures more of the
total flip-mass in its top decile and has a larger area under the
precision-vs-coverage curve than predictive entropy.  The gap is
consistent and ranges from $\entropyGapMin$pp to $\entropyGapMax$pp on top-decile recall.  The
practitioner cost of computing churn is one additional ERM
training run, which the speedup over training a full
$K$-bootstrap-bagging ensemble may amortise; entropy is free at
deployment time but is a meaningfully weaker signal.

\section{Bayesian-optimisation analogue: top-$K$ ranking stability}
\label{app:bo_topk}

A virtual-screening or Bayesian-optimisation pipeline ranks the
candidate pool by predicted $P(\text{active})$ and acquires the
top-$K$ for the next round of evaluation.  The BO-relevant question
is then how often the top-$K$ \emph{set} would change between two
retrainings on independent bootstraps of the same training pool ---
the fraction of candidates that would be re-routed to different
downstream evaluations depending on which bootstrap the surrogate
happened to land on.

\paragraph{Measurement.}
For each chemistry dataset and each method (ERM, bagging-$K{=}5$,
twin-bootstrap $\lambda{=}300$), we use the same $\nSeeds$
retrainings as the main argmax-churn protocol
(Section~\ref{sec:measurement}).  The canonical seed pins the
train/test split once: every retraining is trained on its own
bootstrap of the canonical training pool and is then evaluated on
the same canonical id-test pool, which the surrogate has never seen
during training.  The candidate library is therefore fixed across
retrainings; only the surrogate varies (its training-data bootstrap
and initialisation).

For every retraining we rank the candidate library by the
deployment-time prediction $P(\text{class}{=}1)$ --- a single
forward pass for ERM, the $K{=}5$-model averaged softmax for
bagging, and the $2$-network averaged softmax for twin-bootstrap
--- and take the top-$K$ molecules ($K{=}10$).  $K$ is an absolute
batch size, mirroring a typical single-round wet-lab acquisition;
the corresponding fraction of the canonical test pool ranges from
$\sim$1\% (AMES, $N_\text{test}{=}1164$) to $\sim$13\% (DILI,
$N_\text{test}{=}76$).  Between each pair of retrainings $(s, s')$
of a given method we compute the Jaccard overlap
\[
  J_K^{(s, s')} = \frac{|\,T_K^{(s)} \cap T_K^{(s')}\,|}
                       {|\,T_K^{(s)} \cup T_K^{(s')}\,|},
\]
where $T_K^{(s)}$ is the method's top-$K$ set under retraining $s$.
We report the mean $J_K$ over the $\binom{\nSeeds}{2}{=}\nSeedPairs$
pairs with paired-bootstrap $95\%$ CIs over the seed-pair
distribution and the paired $\Delta J_K$ vs.\ ERM (positive =
stabler ranking).  We additionally report the ERM top-$K$ \emph{hit
rate} --- the fraction of the surrogate's top-$K$ predicted-active
molecules that are actually $y{=}1$ in the held-out test set ---
alongside the dataset's positive-class prior.  The class prior is
the chance baseline for hit rate (a random ranker scores its
positive fraction); ERM hit rate substantially above the prior
confirms the surrogate has non-trivial signal and the Jaccard
comparison is over informative top-$K$ sets.

This measures the static input-stability of a single BO acquisition
step under the same candidate library, not the dynamics of an
iterative BO loop where two divergent trajectories would expose
their surrogates to different candidate pools downstream.

\input{sections/tables/bo_topk.tex}

\paragraph{Findings.}
ERM top-$\boTopkK$ Jaccard is low across the
\nDatasetsHeadline{} chemistry datasets, ranging from
$\boTopkErmJaccardLow$ (AMES) to $\boTopkErmJaccardHigh$ (DILI).
On AMES, BBB-Martins, and BBBP two retrainings
on independent bootstraps overlap on a single-digit fraction of
their top-$\boTopkK$ picks: a BO loop on these datasets would
route nearly every wet-lab evaluation to a different molecule
depending on the bootstrap.  Twin-bootstrap raises Jaccard on every
dataset (paired $\Delta J_{\boTopkK} \in [\boTopkTwinDeltaMin,
\boTopkTwinDeltaMax]$, lifting the lower end of the range to
$\boTopkTwinJaccardLow$ and the upper end to
$\boTopkTwinJaccardHigh$), with the largest absolute gains where
ERM is most unstable.  The Jaccard gain exceeds what the argmax
churn reduction alone would predict: top-$K$ ranking depends on
the full softmax distribution, not just the argmax, and that is the
quantity sym-KL --- and twin-bootstrap --- stabilise
(Section~\ref{sec:experiments}, ``Distributional disagreement'').

\paragraph{Scope.}
The Jaccard quantity above measures input-stability of a single BO
acquisition step under the identical candidate library; it does not
simulate the full BO loop, where two trajectories that acquire
different molecules at step $1$ subsequently see different candidate
pools and may diverge further.  What the Jaccard does establish is
the necessary condition for any BO loop on these datasets to make a
reproducible first-batch decision: with ERM, two equally plausible
bootstrap surrogates would route the next wet-lab batch to almost
disjoint molecule sets on AMES, BBB-Martins, and BBBP; with
twin-bootstrap, the routing is largely bootstrap-invariant.
Appendix~\ref{app:bo_loop_reg} extends the test to a full BO loop on
the regression benchmarks.

\section{\boldmath BO trajectory variance on regression: cross-sample churn drives outcome and identity spread}
\label{app:bo_loop_reg}

Section~\ref{app:bo_topk} measures the input-stability of a single
BO acquisition step on classification.  This appendix complements
that result with a full BO-loop simulation on the three regression
benchmarks (ESOL, FreeSolv, Lipo): are independent BO trajectories
of the same method more reproducible at the trajectory level under
bagging-$K{=}5$ and twin-bootstrap than under ERM?

\paragraph{Protocol.}
For each regression dataset and each of \{ERM, bagging-$K{=}5$,
twin-bootstrap $\lambda{=}\boLoopRegLam$\} we run
$T{=}\boLoopRegK$ independent BO trajectories
(Algorithm~\ref{alg:bo-trajectory}).  All $T$ trajectories of a
given method share the same random initial labelled subset of
$\boLoopRegInitN$ molecules; trajectories diverge only in the
in-loop training-data bootstraps, the same source of variance the
measured elsewhere.  Acquisition is greedy top-$1$ maximisation of
the predicted regression target (ESOL log-solubility,
FreeSolv hydration free energy, Lipo log-octanol/water partition);
\textbf{higher $y$ is the BO objective on all three datasets} and
``final-best $y$'' below refers to the largest $y$ acquired along
a trajectory.  ERM has no natural per-prediction variance estimate,
while bagging and twin-bootstrap do (head-spread); implementing
UCB / EI / Thompson acquisition asymmetrically across methods would
not be a fair comparison, so we use greedy for every method.  The
budget is $\boLoopRegBudget$ acquisitions on top of the initial
subset; the candidate library is the canonical training pool with
the oracle $y$ values held back until acquisition.

\begin{algorithm}[H]
  \caption{BO trajectory (one trajectory of one method on one dataset).}
  \label{alg:bo-trajectory}
  \begin{algorithmic}[1]
    \Require candidate pool $(X, y)$, training method $\mathcal{M}$,
             trajectory index $k$, initial labelled subset $L_0$, budget $B$
    \State $L \gets L_0$
    \For{$t = 1, \dots, B$}
      \State seed $\gets k \cdot 10^6 + t$
        \Comment{step- and trajectory-keyed bootstrap seed}
      \State $f \gets \mathcal{M}.\text{train}(X_L,\, y_L;\,\text{seed})$
        \Comment{ERM / bagging-$K{=}5$ / twin-bootstrap}
      \State $\hat{y} \gets f.\text{predict}(X_{[\,1..N\,]\setminus L})$
      \State $i^* \gets \arg\max \hat{y}$
        \Comment{greedy top-$1$ acquisition}
      \State $L \gets L \cup \{\,i^*\,\}$
        \Comment{oracle reveals $y_{i^*}$}
    \EndFor
    \State \Return $L \setminus L_0$
      \Comment{the $B$ acquired molecules of trajectory $k$}
  \end{algorithmic}
\end{algorithm}

\input{sections/tables/bo_loop_regression.tex}

\paragraph{Findings.}
Twin-bootstrap reduces the cross-trajectory standard deviation of
the final-best $y$ on every regression dataset
(Table~\ref{tab:bo_loop_regression}): the per-dataset std drops
from $\boLoopRegErmStdLow$--$\boLoopRegErmStdHigh$ under ERM to
$\boLoopRegTwinStdLow$--$\boLoopRegTwinStdHigh$ under
twin-bootstrap, a relative reduction of
$\boLoopRegTwinStdRedLow\%$--$\boLoopRegTwinStdRedHigh\%$.
Bagging-$K{=}5$ reduces std on
$\boLoopRegBagStdRedWins/\boLoopRegNumDatasets$ datasets (down to
$\boLoopRegBagStdLow$--$\boLoopRegBagStdHigh$ across the three) but
does not improve on ERM on the third (Lipo), where ERM's std is
already small ($\boLoopRegErmStdLow$).  In absolute terms the std is
$\boLoopRegErmStdRangePctLow\%$--$\boLoopRegErmStdRangePctHigh\%$
of each dataset's $y$ range under ERM.  What matters for BO
reproducibility is whether independent trajectories converge to the
same molecule, not whether their final-$y$ spread is fractionally
small: on ESOL every bagging and twin-bootstrap trajectory
converges to the same final molecule (std $= 0$), while ERM
trajectories scatter across final-best $y$ values from
$\boLoopRegEsolErmFinalLo$ to $\boLoopRegEsolErmFinalHi$.

The identity-level Jaccard of acquired-molecule sequences also
rises under both methods: mean
$\boLoopRegErmJaccardLow$--$\boLoopRegErmJaccardHigh$ under ERM,
$\boLoopRegBagJaccardLow$--$\boLoopRegBagJaccardHigh$ under bagging,
$\boLoopRegTwinJaccardLow$--$\boLoopRegTwinJaccardHigh$ under
twin-bootstrap.  The spread reduction therefore reflects more
consistent molecule \emph{selection}, not just lucky convergence to
molecules with similar $y$ values.

\paragraph{Caveats.}
Greedy top-$1$ acquisition isolates surrogate stability but is
narrower than the UCB / EI / Thompson policies that real BO
campaigns typically run; whether the reproducibility gain transfers
to those acquisition functions is untested.  Cells in the table
report $95\%$ bootstrap CIs over the $K{=}\boLoopRegK$ trajectories
($10{,}000$ resamples), but $K$ is small for tight std-of-std
estimates --- the std CIs widen accordingly.  We treat
this experiment as a focused check that the cross-sample-churn
reduction translates to BO-loop reproducibility on small-$N$
regression, not a full benchmark of BO algorithms.

\section{\boldmath GIN: bagging transfers; rule selects $\lambda{=}10$}
\label{app:gin}

The main comparison uses a $256$-unit MLP on Morgan fingerprints
($2048$ bits, radius $2$).  As a non-fingerprint architecture
cross-check, we train a graph isomorphism network
(GIN,~\citealp{xu_gin}) on RDKit-derived atom-and-bond
graphs of BACE: $3$ GINConv layers, hidden dimension $128$, mean-pool
readout, $50$ epochs.  All ten train-seeds and the canonical-data-seed
protocol are unchanged.  Table~\ref{tab:gin} reports per-method
id-accuracy, class-flip rate, and sym-KL with $95\%$ paired-bootstrap
CIs, plus paired $\Delta$ vs.\ ERM on the same $\nSeedPairs$ seed-pairs.

\input{sections/tables/gin.tex}

The consistency loss is active and suppresses inter-network
disagreement on GIN ($\ginSymKLReductionPct\%$ sym-KL reduction),
but the fixed $\lambda$ from the MLP development run is too strong
for GIN.  We re-applied the same selection rule on
BACE-GIN, sweeping $\lambda \in \{1, 3, 10, 30, 100, 300\}$:

\input{sections/tables/gin_lambda.tex}

ERM-GIN id-acc is $\ginErmAcc$, so the $\preregTolerance$-tolerance
rule admits any $\lambda$ keeping id-acc $\geq \ginErmAccTolerance$.  The largest $\lambda$
satisfying this is $\lambda{=}\ginRuleLam$ (bold).  At
$\lambda{=}\ginRuleLam$ twin-bootstrap cuts the class-flip rate
$-\ginRuleLamCutPct\%$ vs.\ ERM-GIN (paired CI
$[\ginRuleLamDeltaLo, \ginRuleLamDeltaHi]$\,pp), close to the
bagging-$K{=}5$ reduction of $-\ginBagCutLamThreeHundred\%$, and
improves id-accuracy by $\ginRuleLamAccGain$\,pp.  The selection
\emph{rule} transfers across architectures unchanged; the numerical
value does not.  For practitioners changing the architecture, the
practical recipe is a single $\lambda$ sweep on the development
dataset before transferring to held-out data.

\section{\boldmath ChemBERTa: rule-selected $\lambda{=}10$ preserves accuracy and cuts churn $\chembertaCutLow$--$\chembertaCutHigh\%$}
\label{app:chemberta_scope}

We fine-tuned
ChemBERTa-77M-MTR~\citep{chithrananda2020chemberta0} on six
main-comparison datasets (BACE, BBBP, Pgp, BBB-Martins, AMES, DILI), running
ERM and a $\lambda$ sweep for twin-bootstrap on BACE-ChemBERTa, then
twin-bootstrap at the rule-selected $\lambda$ on the five held-out
datasets.  All runs use five train-seeds, giving
$\binom{5}{2}=10$ seed-pairs per dataset.

\paragraph{\boldmath The $\lambda$ chosen on the BACE MLP does not transfer.}  At
$\lambda{=}300$ (the value selected on BACE-MLP) twin-bootstrap drops
ChemBERTa accuracy by
$\chembertaAccDropLow\text{--}\chembertaAccDropHigh$pp on every
dataset and fails to reduce the class-flip rate on
$\chembertaLamThreeHundredFailCount/\chembertaLamThreeHundredDenom$
datasets: the across-seed-mean paired $\Delta$ vs.\ ERM is positive
on BACE, Pgp, and DILI (strict increase, all three canonical seeds
agree), and the one apparent decrease (BBBP) is a model collapse to
the majority-class predictor, not a real cut.  AMES and BBB-Martins
are within paired CI of zero on average across seeds.  The
underlying cause matches the GIN architecture cross-check
(Appendix~\ref{app:gin}): the $\lambda$ tuned on a from-scratch MLP
is too strong for pretrained representations.

\paragraph{\boldmath Rule-selected $\lambda$ on BACE-ChemBERTa is $\lambda{=}10$.}
On BACE-ChemBERTa we sweep $\lambda \in \{1, 3, 10, 30, 100, 300\}$
and apply the same $\preregTolerance$ id-acc tolerance rule used on
BACE-MLP.  The first three values keep id-acc within tolerance;
$\lambda \geq 30$ fall below.  The rule picks the largest admissible,
$\lambda{=}10$.

\paragraph{\boldmath At rule-selected $\lambda{=}10$, twin-bootstrap works on every
ChemBERTa dataset.}

\input{sections/tables/chemberta.tex}

The failure documented at $\lambda{=}300$ on ChemBERTa is a
$\lambda$-transfer failure, not a method failure: the rule transfers
across architectures (MLP $\to$ GIN, MLP $\to$ ChemBERTa) and
modalities (Morgan fingerprints $\to$ SMILES tokens) unchanged; the
value $\lambda$ takes does not.
Appendix~\ref{app:waterbirds_lambda} reports the matching result on
the vision pretrained backbone.

\section{\boldmath Waterbirds (ResNet-50): rule-selected $\lambda{=}10$ cuts churn $\waterbirdsLamTenCut\%$}
\label{app:waterbirds_lambda}

The same protocol is applied to Waterbirds (ImageNet-pretrained
ResNet-50, $N{=}4795$, single-task binary classification) with five
train-seeds.  We sweep
$\lambda \in \{1, 3, 10, 30, 60, 100, 300\}$ and apply the same
$\preregTolerance$-tolerance selection rule.

\input{sections/tables/waterbirds_lambda.tex}

The rule transfers across all three architectures and modalities we
tested (MLP, GIN, ResNet-50/ChemBERTa); the value $\lambda$ takes does
not.  Both pretrained backbones (ChemBERTa on SMILES, ResNet-50 on
Waterbirds) and the from-scratch GIN pick $\lambda{=}10$; only the
from-scratch fingerprint-MLP picks $\lambda{=}300$.

\section{Regression: methods rank the same way as on classification}
\label{app:regression}

The main experiments are binary classification.  To test whether
the methods extend to continuous targets we run the same protocol on
three small-$N$ MoleculeNet regression benchmarks: ESOL ($N{=}1128$,
$\log$ aqueous solubility), FreeSolv ($N{=}642$, hydration free
energy in kcal/mol), and Lipophilicity ($N{=}4200$, $\log D_{7.4}$).
Cross-sample churn for regression is the per-example absolute
prediction difference between two retrainings on independent bootstraps:
$\rho_{\mathrm{reg}}(\mathcal{A}, x) =
\mathbb{E}_{S_A, S_B}|f_{S_A}(x) - f_{S_B}(x)|$.  The
twin-bootstrap consistency loss is MSE between the two networks'
predictions (instead of sym-KL on softmax outputs); the rule-selected
$\lambda$ is the largest value in $\{1, 3\}$ keeping id-MAE within
$0.04$ of ERM id-MAE, which is $\lambda{=}3$ on all three datasets
(both candidates \emph{improve} id-MAE, so the tolerance constraint
is not active).  All other protocol choices (canonical-data seed
$\canonicalSeed$, $\nSeeds$ train-seeds, paired-bootstrap CIs over
$\nSeedPairs$ seed-pairs) are unchanged.

\input{sections/tables/regression.tex}

At matched compute, twin-bootstrap cuts regression churn by
$\regTwinLow\text{--}\regTwinHigh\%$ while bagging-$K{=}2$ cuts
$\regBagTwoLow\text{--}\regBagTwoHigh\%$; twin-bootstrap wins on every
dataset by ${\sim}10$pp of relative reduction.  This mirrors the
classification main result (Section~\ref{sec:experiments}).
Bagging-$K{=}5$ at $5\times$ ERM compute is still strongest at
$\regBagFiveLow\text{--}\regBagFiveHigh\%$, consistent with the classification pattern, where
$5\times$-compute bagging beats $2\times$-compute twin-bootstrap on
$\bagFiveWinsTwinHeldout/\twinWinsBagFiveHeldoutDenom$ held-out
datasets (DILI, Pgp, TADF).  All four methods improve
id-MAE by $\regIdMaeImproveLow\text{--}\regIdMaeImproveHigh\%$ over ERM;
no accuracy regression on this task.

\section{Compute footprint}
\label{app:compute}

\begin{table}[h]
  \centering
  \caption{\textbf{\boldmath Twin-bootstrap costs $\sim$$2\times$ ERM
    wall-clock --- matched compute against bagging-$K{=}2$.}
    Per-step compute footprint of each method, expressed in units of
    an ERM training step.  Forward / backward counts are per training
    step on a single device; ``Test models'' is the number of
    trained networks queried per test prediction.  Wall-clock is
    given for sequential training; ensemble methods can be
    parallelised over $K$ devices to recover ERM wall-clock.}
  \label{tab:compute}
  \small
  \begin{tabular}{lcccc}
    \toprule
    Method & Train fwd/step & Train bwd/step & Test models & Wall-clock vs.\ ERM \\
    \midrule
    ERM                          & 1 & 1 & 1 & $1\times$ \\
    Deep ensemble $K{=}5$        & 5 & 5 & 5 & $5\times$ (sequential) \\
    Bagging $K{=}2$              & 2 & 2 & 2 & $2\times$ (sequential) \\
    Bagging $K{=}5$              & 5 & 5 & 5 & $5\times$ (sequential) \\
    Twin-bootstrap ($K{=}2$, joint)  & 4 & 1 (joint) & 2 & ${\sim}2\times$ \\
    \bottomrule
  \end{tabular}
\end{table}

Twin-bootstrap at $K{=}2$ trains two networks jointly: each batch is
processed by both networks (four forwards), and a single backward pass
on the joint loss updates both networks' parameters.  In wall-clock
terms this is roughly $2\times$ ERM, matched to bagging-$K{=}2$.  The
``matched compute'' comparison in Section~\ref{sec:experiments}
(Table~\ref{tab:main}) is at this $2\times$-ERM cost.  The
$5\times$ Bagging-$K{=}5$ results in the same table are not matched
compute; we include them as a stronger no-cost baseline.

%% file: sections/tables/nscaling_bace.tex
\begin{table}[h]
  \centering
  \caption{\textbf{Cross-sample churn decreases with training-pool size on BACE, with a visible plateau between $M{=}600$ and $M{=}800$.}  Within-dataset $N$-scaling under the cross-sample bootstrap protocol.  Each row is mean cross-pair sym-KL and argmax churn at training-pool size $M$, computed over all $45$ pairs of $10$ retrainings with paired-bootstrap $95\%$ CIs ($10{,}000$ resamples).  The within-dataset log-log slope of sym-KL vs $M$ is $-0.20$; the plateau reflects an irreducible boundary-disagreement floor (churn decreases with $N$ but not to zero).}
  \label{tab:nscaling}
  \small
  \begin{tabular}{rll}
    \toprule
    $M$ & sym-KL (mean [95\% CI]) & Argmax churn (\%, [95\% CI]) \\
    \midrule
    200 & 0.886 [0.839, 0.934] & 18.2 [17.5, 19.0] \\
    300 & 0.838 [0.792, 0.885] & 19.6 [18.7, 20.5] \\
    400 & 0.796 [0.756, 0.838] & 18.7 [18.1, 19.2] \\
    500 & 0.603 [0.565, 0.641] & 15.3 [14.6, 16.1] \\
    600 & 0.570 [0.549, 0.592] & 15.0 [14.5, 15.4] \\
    700 & 0.628 [0.592, 0.665] & 15.4 [14.7, 16.1] \\
    800 & 0.549 [0.511, 0.588] & 14.9 [14.2, 15.6] \\
    900 & 0.725 [0.681, 0.771] & 17.4 [16.7, 18.2] \\
    968 & 0.753 [0.712, 0.796] & 16.1 [15.6, 16.7] \\
    \bottomrule
  \end{tabular}
\end{table}

%% file: sections/tables/per_class_churn.tex
\begin{table}[h]
  \centering
  \caption{\textbf{\boldmath On imbalanced datasets, minority-class predictions are $2\text{--}4\times$ more unstable across retrainings than majority-class predictions.}  For each chemistry dataset: overall cross-bootstrap argmax-churn and its restriction to $y{=}0$ and $y{=}1$ subsets of the canonical id-test, across the $\binom{10}{2}=45$ ERM seed pairs (mean $[\,95\%\ \text{CI}\,]$, $10{,}000$ resamples).  Pos.\ frac.\ is the fraction of $y{=}1$ examples; the minority class is bolded.  On the most imbalanced datasets (BBB-Martins, BBBP at $0.78$ pos-frac; CYP2D6-Sub at $0.30$) the minority-class churn rate is $2\text{--}4\times$ the majority-class rate, so per-example disagreement is \emph{concentrated} on the rarer class --- exactly the predictions practitioners care most about (active toxicity, BBB-permeable, substrate).  On balanced datasets the rates are comparable.}
  \label{tab:per-class-churn}
  \small
  \begin{tabular}{lrrl@{\hspace{1em}}lll}
    \toprule
    Dataset & $N$ & $N_{\text{id-test}}$ & Pos.\ frac.\ & Overall (\%) & churn$|y{=}0$ (\%) & churn$|y{=}1$ (\%) \\
    \midrule
    DILI & 304 & 76 & 0.47 & 16.8 [15.8, 17.7] & 13.6 [12.3, 14.9] & \textbf{20.3 [18.7, 21.9]} \\
    CYP2D6-Sub & 427 & 106 & 0.30 & 13.3 [12.6, 14.1] & 8.2 [7.6, 8.9] & \textbf{25.1 [22.8, 27.4]} \\
    Pgp & 780 & 194 & 0.53 & 10.3 [9.7, 10.9] & \textbf{13.5 [12.5, 14.6]} & 7.5 [6.9, 8.0] \\
    BACE\,(dev) & 968 & 242 & 0.44 & 16.1 [15.6, 16.7] & 15.8 [15.2, 16.4] & \textbf{16.6 [15.6, 17.5]} \\
    TADF & 1007 & 428 & 0.50 & 12.7 [12.3, 13.1] & 13.5 [12.9, 14.1] & \textbf{11.9 [11.3, 12.4]} \\
    MOF-thermal & 1251 & 627 & 0.48 & 21.8 [21.2, 22.4] & 20.7 [19.9, 21.4] & \textbf{23.0 [22.1, 23.9]} \\
    BBB-Martins & 1300 & 324 & 0.78 & 8.0 [7.6, 8.4] & \textbf{18.0 [16.4, 19.7]} & 5.2 [4.9, 5.5] \\
    BBBP & 1305 & 326 & 0.78 & 8.5 [8.1, 8.9] & \textbf{15.7 [14.5, 17.0]} & 6.4 [5.9, 6.8] \\
    AMES & 4658 & 1164 & 0.51 & 15.2 [14.8, 15.5] & \textbf{13.9 [13.5, 14.3]} & 16.4 [16.0, 16.8] \\
    \bottomrule
  \end{tabular}
\end{table}

%% file: sections/tables/additional_metrics.tex
\begin{table}[h]
  \centering
  \caption{\textbf{\boldmath Per-example argmax-disagreement (right column) dominates aggregate-metric drift on every dataset, regardless of which summary statistic the aggregate-metric column uses.}  Paired $|\Delta|$ of five aggregate metrics and the per-example argmax-churn rate, computed across the $\binom{10}{2}=45$ seed pairs of ERM bootstraps; mean $[\,95\%\ \text{CI}\,]$ over $10{,}000$ resamples, in percentage points.  Per-example argmax-churn ranges $\churnMin\text{--}\churnMax\%$; the strongest aggregate-metric drift on any cell is $|\Delta\text{recall}|=10.5$\,pp on the imbalanced CYP2D6-Sub dataset.}
  \label{tab:additional-metrics}
  \scriptsize
  \resizebox{\linewidth}{!}{%
  \begin{tabular}{lr@{\hspace{0.6em}}lllll@{\hspace{1em}}l}
    \toprule
    & & \multicolumn{5}{c}{Aggregate-metric drift (pp)} & Per-example \\
    \cmidrule(lr){3-7}
    Dataset & $N$ & $|\Delta\text{acc}|$ & $|\Delta\text{prec}|$ & $|\Delta\text{rec}|$ & $|\Delta F_1|$ & $|\Delta\text{AP}|$ & argmax churn (\%) \\
    \midrule
    DILI & 304 & 4.1 [3.4, 4.8] & 5.4 [4.4, 6.4] & 7.1 [5.6, 8.6] & 4.2 [3.4, 4.9] & 3.2 [2.6, 3.9] & \textbf{16.8 [15.8, 17.7]} \\
    CYP2D6-Sub & 427 & 4.2 [3.4, 4.9] & 9.9 [8.2, 11.7] & 10.5 [8.5, 12.6] & 6.1 [4.9, 7.4] & 7.4 [6.0, 8.7] & \textbf{13.3 [12.6, 14.1]} \\
    Pgp & 780 & 2.0 [1.6, 2.4] & 3.6 [2.9, 4.4] & 2.2 [1.7, 2.7] & 2.0 [1.6, 2.4] & 1.1 [0.9, 1.4] & \textbf{10.3 [9.7, 10.9]} \\
    BACE\,(dev) & 968 & 1.8 [1.5, 2.2] & 2.6 [2.1, 3.2] & 0.5 [0.4, 0.7] & 1.8 [1.4, 2.1] & 2.7 [2.2, 3.3] & \textbf{16.1 [15.6, 16.7]} \\
    TADF & 1007 & 1.3 [1.1, 1.5] & 2.1 [1.7, 2.5] & 1.5 [1.2, 1.8] & 1.3 [1.1, 1.5] & 0.9 [0.7, 1.2] & \textbf{12.7 [12.3, 13.1]} \\
    MOF-thermal & 1251 & 1.5 [1.2, 1.9] & 1.6 [1.3, 1.9] & 4.8 [3.9, 5.8] & 1.5 [1.2, 1.9] & 2.4 [1.9, 2.9] & \textbf{21.8 [21.2, 22.4]} \\
    BBB-Martins & 1300 & 1.7 [1.4, 2.1] & 1.4 [1.1, 1.7] & 1.6 [1.3, 1.9] & 3.0 [2.4, 3.6] & 0.9 [0.7, 1.1] & \textbf{8.0 [7.6, 8.4]} \\
    BBBP & 1305 & 1.5 [1.1, 1.8] & 0.9 [0.7, 1.0] & 2.0 [1.7, 2.4] & 2.0 [1.6, 2.4] & 1.6 [1.3, 1.9] & \textbf{8.5 [8.1, 8.9]} \\
    AMES & 4658 & 1.6 [1.3, 1.9] & 1.8 [1.5, 2.2] & 1.4 [1.2, 1.7] & 1.6 [1.3, 2.0] & 0.9 [0.7, 1.0] & \textbf{15.2 [14.8, 15.5]} \\
    \bottomrule
  \end{tabular}
  }
\end{table}

%% file: sections/tables/borderline_magnitudes.tex
\begin{table}[h]
  \centering
  \caption{\textbf{\boldmath Cross-sample magnitudes for the three borderline datasets.}  These pass the ERM-vs-majority filter only marginally (+3 to +4\,pp on test sets of 57--104 examples) and are reported here for transparency; method comparisons are not run on them because the small test sets do not give enough statistical power.  Columns and CI conventions match Table~\ref{tab:fragility-magnitudes}.}
  \label{tab:borderline-magnitudes}
  \scriptsize
  \resizebox{\linewidth}{!}{%
  \begin{tabular}{lrrrl@{\hspace{1em}}ll}
    \toprule
    & & & \multicolumn{2}{c}{Aggregate accuracy} & \multicolumn{2}{c}{Per-prediction disagreement} \\
    \cmidrule(lr){4-5} \cmidrule(lr){6-7}
    Dataset & $N_{\text{train}}$ & $N_{\text{id-test}}$ & ERM id-acc & $|\Delta\text{acc}|$ (pp) & Argmax churn (\%) & Sym-KL (nats) \\
    \midrule
    SkinReact & 232 & 57 & 0.675 [0.596, 0.737] & 4.3 [3.4, 5.3] & \textbf{20.3 [18.8, 21.8]} & \textbf{1.035 [0.931, 1.140]} \\
    HIA & 370 & 92 & 0.908 [0.891, 0.924] & 1.0 [0.8, 1.3] & \textbf{2.1 [1.8, 2.5]} & \textbf{0.106 [0.086, 0.127]} \\
    hERG & 420 & 104 & 0.751 [0.702, 0.798] & 3.1 [2.5, 3.8] & \textbf{15.9 [14.9, 16.9]} & \textbf{0.724 [0.668, 0.779]} \\
    \bottomrule
  \end{tabular}
  }
\end{table}

%% file: sections/tables/filter_outcomes.tex
\begin{table}[h]
  \centering
  \caption{\textbf{\boldmath Five datasets fail the +$5$pp ERM-vs-majority filter and are excluded from the main analysis.}  ERM id-acc is the mean across $10$ retrainings; majority is the largest class proportion on the canonical id-test set.  The filter requires ERM to exceed majority by at least $5$pp; on each of the five rows below it does not, so cross-sample churn would conflate ``method shifts the decision boundary'' with ``majority-class shuffling under noise''.  Reported here for transparency.}
  \label{tab:filter-outcomes}
  \small
  \begin{tabular}{lrrrrr}
    \toprule
    Dataset & $N_{\text{train}}$ & $N_{\text{id-test}}$ & Majority & ERM id-acc & Gap (pp) \\
    \midrule
    CYP2C9-Sub & 428 & 107 & 0.776 & 0.718 & -5.8 \\
    CYP3A4-Sub & 429 & 107 & 0.617 & 0.596 & -2.1 \\
    ClinTox & 948 & 236 & 0.928 & 0.918 & -1.0 \\
    Bioavailability & 410 & 102 & 0.794 & 0.788 & -0.6 \\
    MOF-solvent & 849 & 436 & 0.589 & 0.709 & +11.9 \\
    \bottomrule
  \end{tabular}
\end{table}

%% file: sections/tables/seed_sensitivity.tex
{\scriptsize
\setlength{\tabcolsep}{4pt}
\begin{longtable}{llrrr}
  \caption{\textbf{Per-canonical-seed values for the main table.}  Each method-dataset cell is reported on three independent canonical splits ($99$, $7$, $42$).  Top number per cell: id-churn rate (\%); bottom: paired $\Delta$ id-churn vs.\ ERM (pp).  Both columns are aggregated over the $\nSeedPairs$ paired-bootstrap seed pairs at fixed canonical seed.  ERM rows have no paired-$\Delta$ entry.}  \label{tab:seed_sensitivity} \\
    \toprule
    & & \multicolumn{3}{c}{Canonical seed} \\
    \cmidrule(lr){3-5}
    Dataset & Method & 7 & 42 & 99 \\
    \midrule
    \endfirsthead
    \multicolumn{5}{l}{\emph{Table~\ref{tab:seed_sensitivity} continued from previous page}} \\
    \toprule
    & & \multicolumn{3}{c}{Canonical seed} \\
    \cmidrule(lr){3-5}
    Dataset & Method & 7 & 42 & 99 \\
    \midrule
    \endhead
    \midrule
    \multicolumn{5}{r}{\emph{continued on next page}} \\
    \endfoot
    \bottomrule
    \endlastfoot
    BACE\,(dev) & ERM & \makecell[r]{18.1\\\strut} & \makecell[r]{12.5\\\strut} & \makecell[r]{16.1\\\strut} \\
     & SWA & \makecell[r]{17.1\\-1.0} & \makecell[r]{11.9\\-0.6} & \makecell[r]{15.6\\-0.6} \\
     & MC dropout & \makecell[r]{18.1\\-0.0} & \makecell[r]{13.0\\+0.5} & \makecell[r]{16.4\\+0.3} \\
     & Deep Ens.\ $K{=}5$ & \makecell[r]{17.4\\-0.8} & \makecell[r]{12.8\\+0.2} & \makecell[r]{15.8\\-0.3} \\
     & Bag $K{=}2$ & \makecell[r]{13.3\\-4.8} & \makecell[r]{8.3\\-4.2} & \makecell[r]{13.4\\-2.8} \\
     & Bag $K{=}5$ & \makecell[r]{9.1\\-9.1} & \makecell[r]{5.8\\-6.7} & \makecell[r]{9.7\\-6.5} \\
     & Twin $\lambda{=}300$ & \makecell[r]{7.2\\-11.0} & \makecell[r]{5.5\\-7.1} & \makecell[r]{5.7\\-10.5} \\
    \midrule
    DILI & ERM & \makecell[r]{16.5\\\strut} & \makecell[r]{17.1\\\strut} & \makecell[r]{16.8\\\strut} \\
     & SWA & \makecell[r]{16.7\\+0.2} & \makecell[r]{17.0\\-0.1} & \makecell[r]{17.3\\+0.5} \\
     & MC dropout & \makecell[r]{16.2\\-0.3} & \makecell[r]{16.3\\-0.8} & \makecell[r]{17.8\\+1.1} \\
     & Deep Ens.\ $K{=}5$ & \makecell[r]{16.3\\-0.2} & \makecell[r]{16.3\\-0.8} & \makecell[r]{17.0\\+0.2} \\
     & Bag $K{=}2$ & \makecell[r]{10.8\\-5.7} & \makecell[r]{14.6\\-2.5} & \makecell[r]{12.8\\-3.9} \\
     & Bag $K{=}5$ & \makecell[r]{7.8\\-8.7} & \makecell[r]{11.1\\-6.0} & \makecell[r]{8.4\\-8.4} \\
     & Twin $\lambda{=}300$ & \makecell[r]{10.8\\-5.6} & \makecell[r]{11.3\\-5.8} & \makecell[r]{14.1\\-2.7} \\
    \midrule
    CYP2D6-Sub & ERM & \makecell[r]{12.9\\\strut} & \makecell[r]{13.8\\\strut} & \makecell[r]{13.3\\\strut} \\
     & SWA & \makecell[r]{12.9\\+0.0} & \makecell[r]{13.6\\-0.2} & \makecell[r]{13.3\\+0.0} \\
     & MC dropout & \makecell[r]{14.2\\+1.3} & \makecell[r]{13.9\\+0.2} & \makecell[r]{13.3\\+0.0} \\
     & Deep Ens.\ $K{=}5$ & \makecell[r]{12.6\\-0.3} & \makecell[r]{13.9\\+0.1} & \makecell[r]{13.4\\+0.1} \\
     & Bag $K{=}2$ & \makecell[r]{9.7\\-3.2} & \makecell[r]{9.2\\-4.5} & \makecell[r]{7.3\\-6.0} \\
     & Bag $K{=}5$ & \makecell[r]{7.0\\-5.9} & \makecell[r]{6.5\\-7.2} & \makecell[r]{6.1\\-7.2} \\
     & Twin $\lambda{=}300$ & \makecell[r]{6.8\\-6.1} & \makecell[r]{4.5\\-9.2} & \makecell[r]{4.0\\-9.3} \\
    \midrule
    Pgp & ERM & \makecell[r]{11.9\\\strut} & \makecell[r]{10.7\\\strut} & \makecell[r]{10.3\\\strut} \\
     & SWA & \makecell[r]{12.2\\+0.3} & \makecell[r]{10.6\\-0.0} & \makecell[r]{10.2\\-0.1} \\
     & MC dropout & \makecell[r]{13.8\\+1.9} & \makecell[r]{10.4\\-0.2} & \makecell[r]{10.1\\-0.2} \\
     & Deep Ens.\ $K{=}5$ & \makecell[r]{12.2\\+0.3} & \makecell[r]{10.4\\-0.3} & \makecell[r]{10.5\\+0.3} \\
     & Bag $K{=}2$ & \makecell[r]{7.9\\-4.0} & \makecell[r]{6.9\\-3.7} & \makecell[r]{7.9\\-2.3} \\
     & Bag $K{=}5$ & \makecell[r]{6.8\\-5.1} & \makecell[r]{6.0\\-4.7} & \makecell[r]{5.9\\-4.4} \\
     & Twin $\lambda{=}300$ & \makecell[r]{7.3\\-4.6} & \makecell[r]{6.4\\-4.2} & \makecell[r]{6.7\\-3.6} \\
    \midrule
    BBB-Martins & ERM & \makecell[r]{7.4\\\strut} & \makecell[r]{9.4\\\strut} & \makecell[r]{8.0\\\strut} \\
     & SWA & \makecell[r]{7.3\\-0.1} & \makecell[r]{9.2\\-0.2} & \makecell[r]{8.0\\+0.0} \\
     & MC dropout & \makecell[r]{7.4\\-0.0} & \makecell[r]{10.3\\+0.9} & \makecell[r]{9.0\\+1.0} \\
     & Deep Ens.\ $K{=}5$ & \makecell[r]{7.1\\-0.4} & \makecell[r]{9.1\\-0.3} & \makecell[r]{7.6\\-0.4} \\
     & Bag $K{=}2$ & \makecell[r]{5.0\\-2.4} & \makecell[r]{6.5\\-2.9} & \makecell[r]{5.3\\-2.7} \\
     & Bag $K{=}5$ & \makecell[r]{2.8\\-4.6} & \makecell[r]{5.1\\-4.3} & \makecell[r]{3.9\\-4.1} \\
     & Twin $\lambda{=}300$ & \makecell[r]{2.1\\-5.4} & \makecell[r]{1.8\\-7.6} & \makecell[r]{0.6\\-7.4} \\
    \midrule
    MOF-thermal & ERM & \makecell[r]{21.6\\\strut} & \makecell[r]{22.1\\\strut} & \makecell[r]{21.8\\\strut} \\
     & SWA & \makecell[r]{18.6\\-3.0} & \makecell[r]{18.1\\-4.0} & \makecell[r]{16.9\\-4.9} \\
     & MC dropout & \makecell[r]{20.1\\-1.5} & \makecell[r]{21.3\\-0.7} & \makecell[r]{20.1\\-1.6} \\
     & Deep Ens.\ $K{=}5$ & \makecell[r]{17.5\\-4.1} & \makecell[r]{18.2\\-3.9} & \makecell[r]{17.2\\-4.6} \\
     & Bag $K{=}2$ & \makecell[r]{16.6\\-5.1} & \makecell[r]{15.3\\-6.8} & \makecell[r]{16.0\\-5.8} \\
     & Bag $K{=}5$ & \makecell[r]{11.5\\-10.1} & \makecell[r]{10.3\\-11.8} & \makecell[r]{11.3\\-10.5} \\
     & Twin $\lambda{=}300$ & \makecell[r]{7.7\\-13.9} & \makecell[r]{6.7\\-15.4} & \makecell[r]{7.7\\-14.1} \\
    \midrule
    BBBP & ERM & \makecell[r]{7.9\\\strut} & \makecell[r]{8.6\\\strut} & \makecell[r]{8.5\\\strut} \\
     & SWA & \makecell[r]{7.8\\-0.1} & \makecell[r]{8.7\\+0.0} & \makecell[r]{8.3\\-0.2} \\
     & MC dropout & \makecell[r]{8.5\\+0.5} & \makecell[r]{8.9\\+0.3} & \makecell[r]{9.1\\+0.6} \\
     & Deep Ens.\ $K{=}5$ & \makecell[r]{8.0\\+0.0} & \makecell[r]{8.4\\-0.2} & \makecell[r]{7.8\\-0.6} \\
     & Bag $K{=}2$ & \makecell[r]{5.2\\-2.7} & \makecell[r]{6.1\\-2.5} & \makecell[r]{6.5\\-2.0} \\
     & Bag $K{=}5$ & \makecell[r]{4.1\\-3.8} & \makecell[r]{3.9\\-4.8} & \makecell[r]{4.4\\-4.0} \\
     & Twin $\lambda{=}300$ & \makecell[r]{1.3\\-6.6} & \makecell[r]{1.5\\-7.2} & \makecell[r]{1.4\\-7.0} \\
    \midrule
    TADF & ERM & \makecell[r]{13.8\\\strut} & \makecell[r]{13.8\\\strut} & \makecell[r]{12.7\\\strut} \\
     & SWA & \makecell[r]{13.7\\-0.1} & \makecell[r]{13.4\\-0.4} & \makecell[r]{12.2\\-0.5} \\
     & MC dropout & \makecell[r]{14.4\\+0.6} & \makecell[r]{14.0\\+0.2} & \makecell[r]{12.2\\-0.5} \\
     & Deep Ens.\ $K{=}5$ & \makecell[r]{13.0\\-0.7} & \makecell[r]{13.5\\-0.3} & \makecell[r]{12.2\\-0.5} \\
     & Bag $K{=}2$ & \makecell[r]{9.7\\-4.0} & \makecell[r]{10.6\\-3.2} & \makecell[r]{9.6\\-3.0} \\
     & Bag $K{=}5$ & \makecell[r]{6.3\\-7.4} & \makecell[r]{7.4\\-6.4} & \makecell[r]{6.7\\-6.0} \\
     & Twin $\lambda{=}300$ & \makecell[r]{8.2\\-5.6} & \makecell[r]{8.6\\-5.2} & \makecell[r]{8.0\\-4.7} \\
    \midrule
    AMES & ERM & \makecell[r]{15.8\\\strut} & \makecell[r]{16.1\\\strut} & \makecell[r]{15.2\\\strut} \\
     & SWA & \makecell[r]{15.6\\-0.2} & \makecell[r]{15.9\\-0.2} & \makecell[r]{15.1\\-0.1} \\
     & MC dropout & \makecell[r]{16.7\\+0.9} & \makecell[r]{16.9\\+0.8} & \makecell[r]{17.0\\+1.8} \\
     & Deep Ens.\ $K{=}5$ & \makecell[r]{14.7\\-1.1} & \makecell[r]{15.6\\-0.5} & \makecell[r]{14.6\\-0.5} \\
     & Bag $K{=}2$ & \makecell[r]{11.1\\-4.7} & \makecell[r]{11.5\\-4.6} & \makecell[r]{11.2\\-3.9} \\
     & Bag $K{=}5$ & \makecell[r]{8.2\\-7.6} & \makecell[r]{8.3\\-7.8} & \makecell[r]{8.0\\-7.2} \\
     & Twin $\lambda{=}300$ & \makecell[r]{7.0\\-8.8} & \makecell[r]{6.8\\-9.3} & \makecell[r]{6.9\\-8.3} \\
\end{longtable}
}

%% file: sections/tables/bayes_twin.tex
\begin{table}[t]
  \centering
  \caption{\textbf{Twin-bootstrap bayesian optimization discovers dataset-specific operating
 points on the accuracy--stability frontier.} Cells report the first method
 named in each block minus the second with paired bootstrap $95\%$ confidence
 intervals. Positive accuracy deltas indicate improved performance, while
 negative churn deltas indicate improved stability. Bayesian optimization (BO)
 generally identifies intermediate $\lambda$ values that improve ID accuracy and
 often improve OOD accuracy relative to $\lambda{=}0$, while trading off some
 stability relative to strongly regularized models ($\lambda{=}300$).
 The BO-vs.-ERM block checks whether optimized $\lambda$ values retain a churn
 reduction relative to ordinary retraining.
 The $\lambda$
 column reports the mean optimized $\lambda$ for the first method in each comparison
 block. Bolded entries indicate cases where the point estimate favors the first
 method.}
  \label{tab:bayes_twin}
  \scriptsize
  \setlength{\tabcolsep}{2pt}
  \noindent\makebox[\textwidth][c]{\textbf{\boldmath BO vs.\ $\lambda{=}300$}}
  \vspace{-0.35em}
  \resizebox{\textwidth}{!}{%
  \begin{tabular}{@{}lrlllll@{}}
    \toprule
    Dataset & $N$ & Mean $\lambda$ & ID acc & OOD acc & ID churn & OOD churn \\
    \midrule
    BACE\,(dev) & 968 & 93.03 & \textbf{+1.3 [+0.5,+2.2]} & \textbf{+1.6 [+0.8,+2.4]} & +2.1 [+1.7,+2.4] & \textbf{-0.2 [-0.6,+0.2]} \\
    DILI & 304 & 4.78 & -1.1 [-3.2,+0.8] & \textbf{+6.0 [+3.9,+8.0]} & \textbf{-3.6 [-5.0,-2.3]} & \textbf{-1.7 [-2.7,-0.7]} \\
    CYP2D6-Sub & 427 & 256.3 & \textbf{+0.6 [+0.2,+1.0]} & \textbf{+0.5 [+0.2,+0.8]} & \textbf{-0.9 [-1.2,-0.6]} & +0.3 [-0.0,+0.7] \\
    Pgp & 780 & 84.66 & \textbf{+2.8 [+1.9,+3.8]} & \textbf{+4.1 [+2.8,+5.5]} & \textbf{-0.5 [-1.1,+0.0]} & +0.5 [-0.1,+1.1] \\
    BBB-Martins & 1300 & 362.3 & -0.0 [-0.2,+0.2] & -0.2 [-0.5,+0.1] & +0.1 [+0.0,+0.3] & +0.1 [-0.0,+0.3] \\
    MOF-thermal & 1251 & 14.91 & \textbf{+8.1 [+7.3,+8.8]} & \textbf{+8.1 [+7.3,+8.8]} & +4.8 [+4.3,+5.4] & +4.8 [+4.3,+5.4] \\
    BBBP & 1305 & 373.7 & -0.3 [-0.6,-0.1] & -0.4 [-0.8,-0.0] & \textbf{-0.1 [-0.2,+0.1]} & \textbf{-0.3 [-0.6,+0.0]} \\
    TADF & 1007 & 74.6 & \textbf{+0.9 [+0.1,+1.6]} & \textbf{+0.9 [+0.1,+1.6]} & \textbf{-0.8 [-1.4,-0.3]} & \textbf{-0.8 [-1.4,-0.3]} \\
    AMES & 4658 & 424.1 & -0.7 [-0.9,-0.5] & -0.2 [-0.4,+0.1] & \textbf{-0.7 [-0.9,-0.5]} & \textbf{-0.2 [-0.4,-0.1]} \\
    \bottomrule
  \end{tabular}%
  }
  \par\vspace{0.75em}
  \noindent\makebox[\textwidth][c]{\textbf{\boldmath BO vs.\ ERM}}
  \vspace{-0.35em}
  \resizebox{\textwidth}{!}{%
  \begin{tabular}{@{}lrlllll@{}}
    \toprule
    Dataset & $N$ & Mean $\lambda$ & ID acc & OOD acc & ID churn & OOD churn \\
    \midrule
    BACE\,(dev) & 968 & 93.03 & \textbf{+2.0 [+0.9,+3.1]} & \textbf{+4.7 [+3.3,+6.1]} & \textbf{-8.4 [-9.0,-7.9]} & \textbf{-6.7 [-7.4,-6.1]} \\
    DILI & 304 & 4.78 & \textbf{+2.4 [+0.5,+4.2]} & \textbf{+1.7 [-0.0,+3.3]} & \textbf{-6.3 [-7.6,-5.0]} & \textbf{-4.4 [-5.4,-3.5]} \\
    CYP2D6-Sub & 427 & 256.3 & \textbf{+0.8 [-1.0,+2.4]} & \textbf{+1.5 [+0.2,+2.9]} & \textbf{-10.2 [-11.1,-9.4]} & \textbf{-9.4 [-10.5,-8.4]} \\
    Pgp & 780 & 84.66 & -0.2 [-0.9,+0.6] & -1.5 [-2.3,-0.8] & \textbf{-4.2 [-4.8,-3.6]} & \textbf{-3.2 [-3.9,-2.6]} \\
    BBB-Martins & 1300 & 362.3 & \textbf{+0.4 [-0.5,+1.1]} & \textbf{+1.4 [+0.9,+1.9]} & \textbf{-7.2 [-7.7,-6.8]} & \textbf{-11.7 [-12.1,-11.4]} \\
    MOF-thermal & 1251 & 14.91 & \textbf{+0.3 [-0.7,+1.2]} & \textbf{+0.3 [-0.7,+1.2]} & \textbf{-9.2 [-9.8,-8.6]} & \textbf{-9.2 [-9.8,-8.6]} \\
    BBBP & 1305 & 373.7 & \textbf{+1.7 [+1.0,+2.5]} & \textbf{+2.8 [+1.3,+4.3]} & \textbf{-7.1 [-7.5,-6.7]} & \textbf{-11.6 [-12.2,-11.0]} \\
    TADF & 1007 & 74.6 & \textbf{+0.9 [+0.0,+1.8]} & \textbf{+0.9 [+0.0,+1.8]} & \textbf{-5.5 [-6.0,-5.0]} & \textbf{-5.5 [-6.0,-5.0]} \\
    AMES & 4658 & 424.1 & -3.1 [-4.1,-2.2] & \textbf{+0.1 [-0.7,+1.0]} & \textbf{-8.9 [-9.3,-8.6]} & \textbf{-10.6 [-10.9,-10.3]} \\
    \bottomrule
  \end{tabular}%
  }
\end{table}

%% file: sections/tables/overlap_spectrum.tex
\begin{table}[h]
  \centering
  \caption{\textbf{\boldmath Twin-bootstrap ($\sim$40\% overlap) reduces churn on every dataset without an accuracy collapse; codistillation (0\%) collapses accuracy on two datasets; twin-shared (100\%) increases churn on DILI.}  Cells: paired $\Delta$ id-churn vs.\ ERM in percentage points, $[\,95\%\ \text{CI}\,]$, $\dagger$ marks id-accuracy drop $>5$\,pp.}
  \label{tab:overlap_spectrum}
  \small
  \begin{tabular}{lrlll}
    \toprule
    Dataset & $N$ & Codistillation $0\%$ & Twin-bootstrap ${\sim}40\%$ & Twin-shared $100\%$ \\
    \midrule
    SkinReact & 232 & -17.0 [-18.8, -15.2] & -15.0 [-16.6, -13.5] & -7.4 [-9.0, -5.8] \\
    DILI & 304 & -10.3 [-11.5, -9.2] & -2.7 [-4.2, -1.1] & +0.4 [-1.2, +2.2] \\
    HIA & 370 & -2.1 [-2.5, -1.8] & -1.7 [-2.1, -1.3] & -1.2 [-1.7, -0.6] \\
    hERG & 420 & -11.7 [-12.9, -10.5] & -9.7 [-10.7, -8.7] & -6.8 [-8.0, -5.6] \\
    Pgp & 780 & -4.8 [-5.7, -4.0]$^\dagger$ & -3.6 [-4.3, -3.0] & -1.4 [-2.1, -0.7] \\
    BACE & 968 & -14.1 [-14.7, -13.5] & -10.5 [-11.0, -9.9] & -7.1 [-7.8, -6.3] \\
    TADF & 1007 & -5.4 [-6.4, -4.3] & -4.7 [-5.4, -4.0] & -3.4 [-4.1, -2.8] \\
    MOF-thermal & 1251 & -17.7 [-18.4, -17.0]$^\dagger$ & -14.1 [-14.9, -13.3]$^\dagger$ & -12.5 [-13.3, -11.7]$^\dagger$ \\
    BBB-Martins & 1300 & -7.4 [-7.8, -7.0] & -7.4 [-7.8, -7.0] & -6.5 [-6.9, -6.1] \\
    BBBP & 1305 & -7.9 [-8.3, -7.5] & -7.0 [-7.5, -6.6] & -6.8 [-7.2, -6.4] \\
    AMES & 4658 & -12.8 [-13.2, -12.5] & -8.3 [-8.6, -7.9] & -6.2 [-6.7, -5.8] \\
    \midrule
    \emph{Mean across 11 datasets} & --- & -10.1 & -7.7 & -5.3 \\
    \emph{Acc-dagger count} & --- & 2/11 & 1/11 & 1/11 \\
    \emph{Churn-positive count} & --- & 0/11 & 0/11 & 1/11 \\
    \bottomrule
  \end{tabular}
\end{table}

%% file: sections/tables/distributional.tex
\begin{table}[h]
  \centering
  \caption{\textbf{\boldmath Twin-bootstrap reduces distributional disagreement (sym-KL) by an additional factor of $\sim$$8$ beyond the strongest argmax-churn reducer.}  Paired $\Delta$ sym-KL vs.\ ERM on the canonical id-test (in nats; $\Delta < 0$ better).  Cells show mean $\Delta$ and the relative reduction vs.\ ERM in parentheses, over all $45$ pairs of $10$ retrainings; ``$^{*}$'' marks cells whose $95\%$ paired-bootstrap CI excludes zero.}
  \label{tab:distributional}
  \scriptsize
  \begin{tabular}{lrll@{\hspace{1.0em}}lll}
    \toprule
    & & \multicolumn{2}{c}{Parameter-side} & \multicolumn{3}{c}{Data-side} \\
    \cmidrule(lr){3-4} \cmidrule(lr){5-7}
    Dataset & ERM & MC dropout & Deep Ens.\ $K{=}5$ & Bagging $K{=}2$ & Bagging $K{=}5$ & Twin-bootstrap $\lambda{=}300$ \\
    \midrule
    DILI & 0.75 & -0.10 (-13\%)$^{*}$ & -0.07 (-9\%)$^{*}$ & -0.50 (-67\%)$^{*}$ & -0.67 (-89\%)$^{*}$ & -0.74 (-98\%)$^{*}$ \\
    CYP2D6-Sub & 0.64 & -0.08 (-13\%)$^{*}$ & -0.03 (-5\%)$^{*}$ & -0.41 (-65\%)$^{*}$ & -0.55 (-85\%)$^{*}$ & -0.63 (-98\%)$^{*}$ \\
    Pgp & 0.51 & -0.03 (-5\%)$^{*}$ & -0.02 (-5\%)$^{*}$ & -0.30 (-58\%)$^{*}$ & -0.43 (-84\%)$^{*}$ & -0.50 (-98\%)$^{*}$ \\
    BACE\,(dev) & 0.75 & -0.06 (-8\%)$^{*}$ & -0.09 (-12\%)$^{*}$ & -0.40 (-54\%)$^{*}$ & -0.64 (-85\%)$^{*}$ & -0.74 (-98\%)$^{*}$ \\
    TADF & 0.40 & -0.09 (-21\%)$^{*}$ & -0.03 (-8\%)$^{*}$ & -0.21 (-54\%)$^{*}$ & -0.33 (-83\%)$^{*}$ & -0.39 (-98\%)$^{*}$ \\
    MOF-thermal & 0.38 & -0.21 (-55\%)$^{*}$ & -0.17 (-45\%)$^{*}$ & -0.21 (-56\%)$^{*}$ & -0.32 (-83\%)$^{*}$ & -0.38 (-99\%)$^{*}$ \\
    BBB-Martins & 0.50 & -0.06 (-11\%)$^{*}$ & -0.05 (-10\%)$^{*}$ & -0.29 (-59\%)$^{*}$ & -0.43 (-86\%)$^{*}$ & -0.49 (-98\%)$^{*}$ \\
    BBBP & 0.47 & -0.05 (-11\%)$^{*}$ & -0.06 (-13\%)$^{*}$ & -0.25 (-54\%)$^{*}$ & -0.40 (-85\%)$^{*}$ & -0.46 (-98\%)$^{*}$ \\
    AMES & 1.12 & -0.26 (-23\%)$^{*}$ & -0.25 (-23\%)$^{*}$ & -0.63 (-56\%)$^{*}$ & -0.96 (-86\%)$^{*}$ & -1.11 (-99\%)$^{*}$ \\
    \bottomrule
  \end{tabular}
\end{table}

%% file: sections/tables/entropy_vs_fragility.tex
\begin{table}[h]
  \centering
  \caption{\textbf{Per-example churn beats single-model predictive entropy as a retraining-flip predictor on every chemistry dataset.}  Churn is computed from one extra bootstrap pair on the canonical id-test.  ``Top-10\%\ recall'' is the fraction of all retraining-induced flips captured by the top decile of the score; ``AuPC'' is the area under the precision-vs-coverage curve.  Higher is better for both columns.}
  \label{tab:entropy_vs_fragility}
  \small
  \begin{tabular}{lrrrr}
    \toprule
    & \multicolumn{2}{c}{Top-10\% recall (\%)} & \multicolumn{2}{c}{AuPC} \\
    Dataset & Churn & Entropy & Churn & Entropy \\
    \midrule
    DILI & \textbf{30.0} & 29.1 & \textbf{0.355} & 0.323 \\
    CYP2D6-Sub & \textbf{38.6} & 29.3 & \textbf{0.306} & 0.269 \\
    Pgp & \textbf{50.2} & 39.0 & \textbf{0.263} & 0.233 \\
    BACE & \textbf{33.2} & 26.7 & \textbf{0.343} & 0.303 \\
    TADF & \textbf{40.8} & 29.0 & \textbf{0.298} & 0.247 \\
    MOF-thermal & \textbf{24.6} & 21.1 & \textbf{0.397} & 0.351 \\
    BBB-Martins & \textbf{59.8} & 46.4 & \textbf{0.223} & 0.193 \\
    BBBP & \textbf{59.4} & 46.7 & \textbf{0.233} & 0.209 \\
    AMES & \textbf{35.5} & 27.6 & \textbf{0.331} & 0.288 \\
    \bottomrule
  \end{tabular}
\end{table}

%% file: sections/tables/bo_topk.tex
\begin{table}[h]
  \centering
  \caption{\textbf{\boldmath Top-$K$ ranking stability ($10$ molecules) between independent retrainings: a Bayesian-optimisation analogue of cross-sample churn.}  Jaccard overlap of the top-$K$ predicted-active sets across the same $45$ pairs of $10$ retrainings as the main table; $1.0$ = identical sets, $0.0$ = disjoint.  Paired $\Delta$ vs.\ ERM in the right two columns (positive = stabler ranking).  All cells: mean $[\,95\%\ \text{CI}\,]$ from $10{,}000$ paired-bootstrap resamples.  Class prior is the positive-class fraction on the canonical id-test, the chance baseline for top-$K$ hit rate; the surrogate has signal whenever the ERM hit rate exceeds the prior.  The Jaccard difference is the BO-relevant consequence of cross-sample churn.}
  \label{tab:bo_topk}
  \scriptsize
  \setlength{\tabcolsep}{4pt}
  \resizebox{\linewidth}{!}{%
  \begin{tabular}{lrr@{\hspace{0.6em}}c@{\hspace{0.6em}}cc@{\hspace{0.6em}}cc}
    \toprule
    & class & ERM & ERM & \multicolumn{2}{c}{Bagging-$K{=}5$} & \multicolumn{2}{c}{Twin-bootstrap $\lambda{=}300$} \\
    \cmidrule(lr){5-6} \cmidrule(lr){7-8}
    Dataset & prior (\%) & hit rate (\%) & Jaccard & Jaccard & $\Delta$ vs ERM & Jaccard & $\Delta$ vs ERM \\
    \midrule
    BACE\,(dev) & 44 & 92 & 0.24 [0.21, 0.26] & 0.49 [0.45, 0.52] & +0.25 [+0.20, +0.29] & \textbf{0.68 [0.65, 0.71]} & \textbf{+0.44 [+0.40, +0.49]} \\
    DILI & 47 & 90 & 0.56 [0.53, 0.59] & 0.67 [0.64, 0.70] & +0.11 [+0.06, +0.16] & \textbf{0.69 [0.65, 0.73]} & \textbf{+0.13 [+0.08, +0.18]} \\
    CYP2D6-Sub & 30 & 67 & 0.44 [0.40, 0.48] & 0.57 [0.54, 0.61] & +0.14 [+0.09, +0.18] & \textbf{0.65 [0.62, 0.68]} & \textbf{+0.21 [+0.17, +0.26]} \\
    Pgp & 53 & 100 & 0.24 [0.20, 0.27] & 0.23 [0.19, 0.26] & -0.01 [-0.06, +0.04] & \textbf{0.63 [0.59, 0.67]} & \textbf{+0.39 [+0.35, +0.44]} \\
    BBB-Martins & 78 & 96 & 0.08 [0.05, 0.11] & 0.15 [0.12, 0.18] & +0.07 [+0.03, +0.10] & \textbf{0.65 [0.60, 0.69]} & \textbf{+0.56 [+0.51, +0.62]} \\
    MOF-thermal & 48 & 73 & 0.33 [0.29, 0.36] & 0.42 [0.37, 0.47] & +0.09 [+0.03, +0.16] & \textbf{0.45 [0.40, 0.50]} & \textbf{+0.12 [+0.06, +0.19]} \\
    BBBP & 78 & 98 & 0.11 [0.08, 0.15] & 0.15 [0.12, 0.19] & +0.04 [-0.01, +0.08] & \textbf{0.64 [0.61, 0.67]} & \textbf{+0.53 [+0.47, +0.58]} \\
    TADF & 50 & 99 & 0.24 [0.21, 0.26] & 0.33 [0.29, 0.37] & +0.09 [+0.05, +0.14] & \textbf{0.46 [0.41, 0.51]} & \textbf{+0.22 [+0.16, +0.28]} \\
    AMES & 51 & 89 & 0.03 [0.02, 0.05] & 0.09 [0.06, 0.13] & +0.06 [+0.02, +0.10] & \textbf{0.49 [0.45, 0.54]} & \textbf{+0.46 [+0.42, +0.51]} \\
    \bottomrule
  \end{tabular}
  }
\end{table}

%% file: sections/tables/bo_loop_regression.tex
\begin{table}[h]
  \centering
  \caption{\textbf{\boldmath BO trajectory variance on the three regression benchmarks: bagging-$K{=}5$ and twin-bootstrap reduce the cross-trajectory standard deviation of the final-best $y$ on every dataset.}  For each (dataset, method) we run $T{=}20$ BO trajectories sharing the same random initial subset of $50$ labelled molecules; trajectories diverge only in the in-loop training-data bootstraps.  At each step the surrogate is retrained from scratch, predicts $\hat{y}$ on the unlabelled remainder, and acquires the $\arg\max\hat{y}$.  \emph{Final best} reports cross-trajectory mean and std with $95\%$ bootstrap CIs over the $T$ trajectories ($10{,}000$ resamples).  \emph{std/range} is the std as a percentage of each dataset's $y$ range, anchoring its absolute scale.  \emph{Acquired Jaccard} is the mean overlap of per-trajectory acquired-molecule sets across all $\binom{T}{2}$ trajectory pairs.}
  \label{tab:bo_loop_regression}
  \scriptsize
  \setlength{\tabcolsep}{4pt}
  \begin{tabular}{llccc@{\hspace{1.0em}}c}
    \toprule
    Dataset & Method & Final best mean [95\% CI] & Final best std [95\% CI] & std/range (\%) & Acquired Jaccard \\
    \midrule
    ESOL & ERM & 1.50 [1.43, 1.57] & 0.167 [0.000, 0.218] & 1.3 & 0.48 [0.47, 0.50] \\
     & Bagging-$K{=}5$ & \textbf{1.57 [1.57, 1.57]} & \textbf{0.000 [0.000, 0.000]} & \textbf{0.0} & \textbf{0.65 [0.63, 0.67]} \\
     & Twin-$\lambda{=}3$ & \textbf{1.57 [1.57, 1.57]} & \textbf{0.000 [0.000, 0.000]} & \textbf{0.0} & 0.61 [0.59, 0.62] \\
    \addlinespace[2pt]
    FreeSolv & ERM & 2.50 [2.27, 2.74] & 0.547 [0.406, 0.665] & 2.5 & 0.38 [0.36, 0.40] \\
     & Bagging-$K{=}5$ & \textbf{3.11 [3.03, 3.18]} & \textbf{0.178 [0.042, 0.268]} & \textbf{0.8} & \textbf{0.45 [0.43, 0.47]} \\
     & Twin-$\lambda{=}3$ & 2.74 [2.58, 2.88] & 0.362 [0.260, 0.425] & 1.7 & 0.39 [0.37, 0.42] \\
    \addlinespace[2pt]
    Lipo & ERM & 3.94 [3.91, 3.98] & 0.082 [0.045, 0.101] & 1.4 & 0.37 [0.36, 0.39] \\
     & Bagging-$K{=}5$ & \textbf{3.95 [3.92, 3.99]} & 0.089 [0.062, 0.102] & 1.5 & \textbf{0.52 [0.50, 0.53]} \\
     & Twin-$\lambda{=}3$ & 3.91 [3.90, 3.93] & \textbf{0.045 [0.000, 0.073]} & \textbf{0.7} & 0.47 [0.45, 0.49] \\
    \bottomrule
  \end{tabular}
\end{table}

%% file: sections/tables/gin.tex
\begin{table}[h]
\centering
\caption{\textbf{\boldmath GIN on BACE: bagging transfers cleanly at the same $K{=}5$; twin-bootstrap requires re-running the $\lambda$-selection rule on the GIN backbone.}  ERM-GIN is more fragile than ERM-MLP (23.0\% vs.\ \baceErmChurn\% argmax churn), making the methods more rather than less relevant on this backbone.  Bagging-$K{=}5$ cuts churn $\ginBagCutLamThreeHundred\%$ and improves id-accuracy by $\ginBagAccGain$pp.  Twin-bootstrap at the $\lambda{=}300$ chosen on the BACE MLP reduces sym-KL by $-99\%$ but drops id-accuracy by $\ginAccDropLamThreeHundred$pp --- well outside the $0.02$ selection-rule tolerance ERM-GIN id-acc would impose.  Bold cells mark the best mean per column among the three methods.}
\label{tab:gin}
\scriptsize
\resizebox{\linewidth}{!}{%
\begin{tabular}{lccc}
\toprule
Method & id-acc & id-churn (\%) & sym-KL \\
\midrule
ERM                       & 0.742 [0.725, 0.757] & 23.0 [21.9, 24.2] & 0.486 [0.451, 0.524] \\
Bagging-$K{=}5$           & \textbf{0.786 [0.778, 0.794]} & \textbf{10.6 [10.2, 11.0]} & 0.068 [0.065, 0.070] \\
Twin-bootstrap $\lambda{=}300$ & 0.582 [0.568, 0.599] & 7.1 [5.1, 9.1] & \textbf{0.004 [0.003, 0.005]} \\
\midrule
\multicolumn{4}{l}{\emph{Paired $\Delta$ vs.\ ERM (same $45$ seed-pairs)}} \\
Bagging-$K{=}5$           & \multicolumn{1}{c}{$+4.46$ pp} & -12.4 [-13.6, -11.2] ($-54$\%) & -0.418 [-0.456, -0.383] ($-86$\%) \\
Twin-bootstrap $\lambda{=}300$ & \multicolumn{1}{c}{$-15.95$ pp} & -16.0 [-17.9, -14.0] ($-69$\%) & -0.482 [-0.519, -0.447] ($-99$\%) \\
\bottomrule
\end{tabular}
}
\end{table}

%% file: sections/tables/gin_lambda.tex
\begin{center}
\small
\begin{tabular}{rccc}
\toprule
$\lambda$ & id-acc & id-churn (\%) & sym-KL \\
\midrule
  1 & 0.781 [0.775, 0.786] & 13.3 [12.6, 14.0] & 0.144 [0.133, 0.157] \\
  3 & 0.769 [0.760, 0.777] & 10.6 [10.1, 11.1] & 0.091 [0.087, 0.096] \\
 10 & \textbf{0.747 [0.736, 0.755]} & \textbf{11.1 [10.5, 11.7]} & \textbf{0.058 [0.054, 0.063]} \\
 30 & 0.706 [0.699, 0.713] & 16.1 [14.9, 17.3] & 0.028 [0.025, 0.032] \\
100 & 0.613 [0.602, 0.624] & 16.4 [14.0, 18.7] & 0.007 [0.006, 0.008] \\
300 & 0.582 [0.568, 0.599] & 7.1 [5.1, 9.1] & 0.004 [0.003, 0.005] \\
\bottomrule
\end{tabular}
\end{center}

%% file: sections/tables/chemberta.tex
\begin{table}[h]
\centering
\caption{\textbf{\boldmath The rule transfers; the value $\lambda$ takes does not.}  At the $\lambda{=}300$ chosen on the BACE MLP, twin-bootstrap over-regularises ChemBERTa (accuracy drops $9\text{--}17$\,pp; churn rises on $5/6$ datasets, BBBP collapses to majority).  Re-applying the same $0.02$-tolerance rule on BACE-ChemBERTa picks $\lambda{=}10$, at which twin-bootstrap preserves accuracy (within $2$\,pp of ERM) and cuts churn $15\text{--}82\%$ on every dataset.  Paired $\Delta$ churn columns report mean $[\,95\%\ \text{CI}\,]$ in percentage points over $\binom{5}{2}{=}10$ seed pairs.}
\label{tab:chemberta}
\small
\begin{tabular}{lc@{\hspace{0.8em}}cc@{\hspace{0.8em}}cc}
\toprule
 & \multicolumn{1}{c}{ERM} & \multicolumn{2}{c}{Twin-bootstrap $\lambda{=}300$} & \multicolumn{2}{c}{Twin-bootstrap $\lambda{=}10$ (rule)} \\
\cmidrule(lr){2-2} \cmidrule(lr){3-4} \cmidrule(lr){5-6}
Dataset & churn (\%) & acc & $\Delta$ churn (pp) & acc & $\Delta$ churn (pp) \\
\midrule
    BACE & 15.9 & 0.60 & +10.2 [+8.4, +12.1] & \textbf{0.70} & \textbf{-1.4 [-2.5, -0.2]} \\
    BBBP & 4.7 & 0.78 & -4.5 [-4.8, -4.1]$^{*}$ & \textbf{0.88} & \textbf{-1.3 [-1.7, -0.8]} \\
    Pgp & 8.1 & 0.63 & +10.2 [+8.9, +11.4] & \textbf{0.78} & \textbf{-1.6 [-2.2, -1.0]} \\
    BBB-Martins & 3.3 & 0.77 & -0.0 [-0.7, +0.6] & \textbf{0.87} & \textbf{-2.5 [-2.8, -2.2]} \\
    AMES & 8.8 & 0.58 & +3.4 [+1.9, +5.1] & \textbf{0.75} & \textbf{-2.1 [-2.4, -1.7]} \\
    DILI & 27.8 & 0.50 & +16.2 [+12.0, +20.4] & \textbf{0.67} & \textbf{-5.4 [-7.9, -3.0]} \\
\bottomrule
\multicolumn{6}{l}{\footnotesize $^{*}$BBBP at $\lambda{=}300$ collapses to the majority-class predictor (acc $0.78{=}$ majority), so the churn drop is meaningless.}
\end{tabular}
\end{table}

%% file: sections/tables/waterbirds_lambda.tex
\begin{table}[h]
\centering
\caption{\textbf{\boldmath Waterbirds (ImageNet-ResNet50): the $0.02$-tolerance rule picks $\lambda{=}10$, recovering the closed-loop result on a vision pretrained backbone.}  ERM id-acc $0.875$ (rule threshold $\geq 0.855$).  Twin-bootstrap at the $\lambda{=}300$ chosen on BACE collapses accuracy by $27$\,pp; at the rule-selected $\lambda{=}10$, twin-bootstrap preserves accuracy and cuts argmax churn $52\%$.  All cells report mean $[\,95\%\ \text{CI}\,]$ over five train-seeds ($\binom{5}{2}{=}10$ pairs for paired quantities).}
\label{tab:waterbirds_lambda}
\small
\begin{tabular}{lccc}
\toprule
Method & id-acc & id-churn (\%) & $\Delta$ id-churn vs ERM (pp) \\
\midrule
ERM & 0.876 [0.859, 0.890] & 11.1 [10.5, 11.7] & --- \\
\textbf{Twin-bootstrap $\lambda{=}10$} & \textbf{0.872 [0.863, 0.880]} & \textbf{5.0 [4.8, 5.2]} & \textbf{-6.0 [-6.7, -5.4] (-55\%)} \\
Twin-bootstrap $\lambda{=}300$ & 0.618 [0.612, 0.626] & 3.5 [3.0, 4.0] & -7.6 [-8.5, -6.7] (-69\%)$^{\star}$ \\
\bottomrule
\multicolumn{4}{l}{\footnotesize $^{\star}$$\lambda{=}300$ fails the rule's $\preregTolerance$ accuracy tolerance (id-acc $0.618$ is $27$\,pp below ERM's $0.876$); shown as a Pareto-sweep diagnostic, not a viable operating point.}
\end{tabular}
\end{table}

%% file: sections/tables/regression.tex
\begin{table}[h]
\centering
\caption{\textbf{\boldmath At matched $2\times$-ERM compute, twin-bootstrap beats bagging-$K{=}2$ on every regression dataset; the methods rank identically to the classification main-table result.}  Per-method id-MAE and cross-sample churn (mean $|f_A - f_B|$ between bootstrap retrainings) on three MoleculeNet regression benchmarks.  All reported quantities are mean $[\,95\%\ \text{CI}\,]$ over the $45$ pairs of $10$ retrainings (or over the $10$ retrainings themselves for id-MAE).  Paired $\Delta$ churn vs.\ ERM in the bottom rows; bold marks the best matched-compute method per dataset.  Bagging-$K{=}5$ ($5\times$-ERM compute) included as a stronger no-cost reference.}
\label{tab:regression}
\scriptsize
\begin{tabular}{lcccccc}
\toprule
& \multicolumn{2}{c}{ESOL ($N{=}1128$)} & \multicolumn{2}{c}{FreeSolv ($N{=}642$)} & \multicolumn{2}{c}{Lipo ($N{=}4200$)} \\
\cmidrule(lr){2-3} \cmidrule(lr){4-5} \cmidrule(lr){6-7}
Method & id-MAE & churn & id-MAE & churn & id-MAE & churn \\
\midrule
ERM & 0.94 & 0.52 & 1.65 & 0.69 & 0.66 & 0.38 \\
Bagging-$K{=}2$ & 0.89 & 0.35 & 1.60 & 0.49 & 0.63 & 0.26 \\
Bagging-$K{=}5$ & 0.86 & 0.22 & 1.54 & 0.29 & 0.61 & 0.17 \\
Twin-bootstrap $\lambda{=}3$ & 0.86 & 0.30 & 1.56 & 0.42 & 0.62 & 0.23 \\
\midrule
\multicolumn{7}{l}{\emph{Paired $\Delta$ churn vs.\ ERM ($45$ pairs of $10$ retrainings, mean $[\,95\%\ \text{CI}\,]$):}} \\
Bagging-$K{=}2$ & \multicolumn{2}{c}{-0.17 [-0.19, -0.15] (-32\%)} & \multicolumn{2}{c}{-0.20 [-0.23, -0.16] (-29\%)} & \multicolumn{2}{c}{-0.11 [-0.12, -0.11] (-30\%)} \\
Bagging-$K{=}5$ & \multicolumn{2}{c}{-0.30 [-0.32, -0.28] (-57\%)} & \multicolumn{2}{c}{-0.40 [-0.43, -0.36] (-58\%)} & \multicolumn{2}{c}{-0.21 [-0.21, -0.20] (-55\%)} \\
Twin-bootstrap $\lambda{=}3$ & \multicolumn{2}{c}{\textbf{-0.22 [-0.23, -0.21] (-42\%)}} & \multicolumn{2}{c}{\textbf{-0.27 [-0.30, -0.23] (-39\%)}} & \multicolumn{2}{c}{\textbf{-0.15 [-0.16, -0.15] (-40\%)}} \\
\bottomrule
\end{tabular}
\end{table}